\documentclass{article}

\usepackage{arxiv}

\usepackage[utf8]{inputenc} % allow utf-8 input
\usepackage[T1]{fontenc}    % use 8-bit T1 fonts
\usepackage{hyperref}       % hyperlinks
\usepackage{url}            % simple URL typesetting
\usepackage{booktabs}       % professional-quality tables
\usepackage{amsfonts}       % blackboard math symbols
\usepackage{nicefrac}       % compact symbols for 1/2, etc.
\usepackage{microtype}      % microtypography
\usepackage{lipsum}		% Can be removed after putting your text content
\usepackage{graphicx}
\usepackage{natbib}
\usepackage{doi}
\usepackage{caption}
\usepackage{subcaption}
\usepackage{algorithm}
\usepackage{algorithmic}

\title{Moments for Perceptive Narration Analysis Through the Audience’s Emotional Attachment\\ to Discourse and Story}

\date{October 5, 2023}	% Hard code date per wishes of Arxiv
\author{  \href{https://orcid.org/0009-0008-8109-2686}{\includegraphics[scale=0.06]{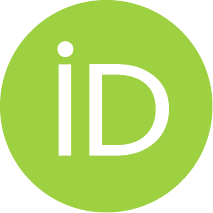}\hspace{1mm}Gary Bruins}\\
School of Performance, Visualization and Fine Arts
	\\ Texas A\&M University, College Station, TX, 77831\\
	\texttt{garybruins@hotmail.com} \\
 \And
 \href{https://orcid.org/0000-0003-3618-4166}{\includegraphics[scale=0.06]{orcid.pdf}\hspace{1mm}Ergun Akleman}\thanks{Joint with Computer Science and Engineering Department.} \\
	Visual Computing \& Computational Media,\\ Texas A\&M University, College Station, TX, 77831\\
	\texttt{ergun@tamu.edu} \\
}

\renewcommand{\shorttitle}{Moments for Storytelling}

\hypersetup{
pdftitle={A template for the arxiv style},
pdfsubject={q-bio.NC, q-bio.QM},
pdfauthor={David S.~Hippocampus, Elias D.~Striatum},
pdfkeywords={First keyword, Second keyword, More},
}

\begin{document}
\maketitle

\begin{abstract}
In this work, our goal is to develop a theoretical framework that can eventually be used for analyzing the effectiveness of visual stories such as feature films to comic books. To develop this theoretical framework, we introduce a new story element called moments. Our conjecture is that any linear story such as the story of a feature film can be decomposed into a set of moments that follow each other. Moments are defined as the perception of the actions, interactions, and expressions of all characters or a single character during a given time period. We categorize the moments into two major types: "story moments" and "discourse moments." Each type of moment can further be classified into three types, which we call universal storytelling moments. We believe these universal moments foster or deteriorate the audience’s emotional attachment to a particular character or the story. We present a methodology to catalog the occurrences of these universal moments as they are found in the story. The cataloged moments can be represented using curves or color strips. Therefore, we can visualize a character's journey through the story as either a 3D curve or a color strip. We also demonstrated that both story and discourse moments can be transformed into one lump-sum ``attraction'' parameter. 
The attraction parameter in time provides a function that can be plotted graphically onto a timeline illustrating changes in the audience’s emotional attachment to a character or the story. By inspecting these functions the story analyst can analytically decipher the moments in the story where the attachment is being established, maintained, strengthened, or conversely where it is languishing. 
 
\end{abstract}

\begin{figure}[htbp!]
 \includegraphics[width=1.0\linewidth]{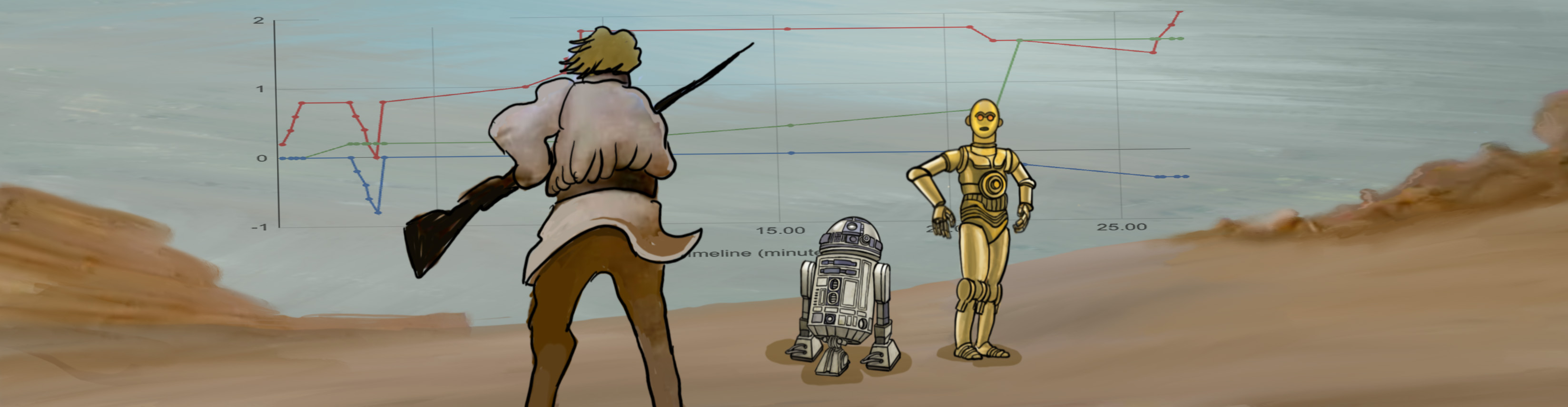}
 \centering
  \caption{A visual interpretation of a moment from a Star Wars movie illustrated by one of the authors. We turn moments such as this into functions that can allow us to analyze the stories (see the function drawn in the sky). The illustration is done by one of the authors. }
\label{fig:teaser}
\end{figure}

\section{Introduction and Motivation}

Narratological analysis is important for a wide variety of applications. For instance, for platforms like Netflix or Amazon, such analysis can be used to predict a particular audience's preferences and improve the suggestion process. Effective analysis can also help to make popular movies that are particularly targeted to large groups of people with similar tastes. Since the major portion of any movie production process is spent on the development of stories and storyboards, with effective analysis the problems in storytelling can be identified early on by speeding the production process. 

Unfortunately, existing models are not that useful for practical purposes. For instance, there has been a significant amount of literature to represent linear stories as a series of cause-and-effect relationships that are constructed in some specific universal structures, which are called plots \cite{forster2010aspects, Barthes1966, Lakoff1972, Greimas1983, tobias1999, Lakoff2010, Akleman2012, caldwell2017story}. These representations are extremely powerful in constructing almost all aspects of storytelling \cite{akleman2015theoretical}. However, they are geared to representation, not to analysis. 
The main problem with all these approaches is that they are too complicated to use in the perceptive analysis. 

Such representations require the use of extended versions of directed acyclic graphs (DAG) of causal inference theory \cite{pearl1995, pearl2000, akleman2015theoretical}. This representation is very powerful in analyzing specific social \cite{pearl1988,pearl1986} and specific economic interactions \cite{bessler1998,akleman1999}. However, storytelling cannot be confined to any specific case. For instance, in movies dead people such as Zombies and Ghosts can exist, and people can move in time. The general nature complicates story representation in such a way that each vertex of DAG can also be a general graph in a recursive way \cite{akleman2015theoretical}. There is, therefore, a need for a simpler representation that can effectively ignore complexities in stories by focusing on only perceptual aspects.

\subsection{Context \& Motivation}

For narratological analysis, there is a need for intermediate approaches that can be sufficiently simplified while still capturing the structures of the stories. In this paper, we present a new approach for representing stories. 

In our approach, we assume a story consists of shorter story segments of a given length we call moments. We restrict our interest to only moments of particular types, more specifically, universal storytelling moments that foster or deteriorate the audience’s emotional attachment to the story or a specific character. We define emotional attachment as the degree to which the audience is attracted, indifferent, or repulsed by a character or story. Note that the story segments are still complicated since they will still need to be represented as stories, which should normally be represented as directed acyclic graphs (DAG) where each vertex can also be a general graph. The major advantage of the moments, they focus on only the perception of either a specific character or the story, evoked by a story segment in the given period of time. Using this simplification, we can ignore complex graph structures and view the story as a few ordered lists of universal moments.   

This simplification can be viewed as analogous to piecewise linear approximations of the non-linear high dimensional functions. Our approach can also be viewed as reminiscent of Riemann's view of the perceptual spaces  \cite{riemann1868hypothesen}. Riemann and later Helmholtz and Schrodinger further postulated that perceptual color space, in particular, is a Rimennian 3-manifold, i.e. it has a shape that locally resembles 3D Euclidean space near each point and it is smooth everywhere  \cite{riemann1868hypothesen, schrodinger1920grundlinien}. The scientific community built perceptive color systems such as CIElab or CIEXYZ based on this underlying principle. The Riemaniann nature of the perceptive color space is recently questioned, but Perceptive color spaces are still viewed as 3D manifolds \cite{bujack2022non}. 

We want to point out that if we ignore perception, the color spaces must require much higher dimensions. On the other hand, as soon as we include humans in the process we can reduce the high dimension to three since humans usually have three cones. This analogy can provide an approach for simplifying narrative structures. Two main elements of narrative structures, discourse, and story form high-dimensional structures. Instead of focusing on actual discourse and stories, we shift our attention to the audience's perception. This shift of focus allows us to significantly reduce the dimension of the problem. 

We observe that perceptual spaces for discourse and story can essentially be simplified into compact 3-manifolds (like colors) that are embedded in high dimensions. Even if they are not Riemennian manifolds, we expect, they are at least reasonably smooth such that we can have reasonably acceptable neighbors similar to color spaces. We expect that these structures may not necessarily be connected. In other words, there can be sufficient diversity of audience types such that these spaces may consist of more than one 3-manifold. To develop such a theoretical framework for narratology, we first need to convert each story into a point in a high-dimensional perceptive space. In this paper, we mainly focus on this problem. 

\subsection{Basis and Rationale}\label{sec:basis}
\label{BasisAndRationale}
The key insight in this paper is that the emotional attachment of the audience to a specific character or a story is essentially a type of perception. Although the story segment can be complicated, the potential perceptive cases can be categorized into a small number of distinct types of moments based on what kind of emotional attachment they evoke in the audience. %We also want to point out that emotional attachments are not emotions. Although there is only a limited number of studies on evoked emotions in Psychology literature they generally support our observation} \cite{scheier1977self, ben1993envy, di2019curiosity. 
By following structuralist terminology \cite{sturrock2008structuralism}, we observe that there is a need for two primary categories of moments: (1) discourse, and (2) story. Discourse-type moments pertain to the audience's emotional attachments to how the story is told, and Story-type moments pertain to the audience's emotional attachments to what is told.

The key observation in this paper is that each of the two types of moments evokes "mainly" three types of emotions. Therefore, each of the two types of moments can further be classified into three sub-types, which we call universal storytelling moments.
Each of these types of universal moments can be considered a linearly independent axis between $1$ and $-1$, where one represents positive emotion and minus one represents negative emotion. Let $\mathbf{m}_{D}$ represent a Discourse moment, $\mathbf{m}_{D} \in [-1,1]^3$, i.e. the moment will be a 3D vector in a cubical domain, likewise for a Story moment.

For the construction of perceptual discourse and story spaces, there is a need for large-scale surveys collected by streaming companies. Since there are only three independent moments, these spaces will essentially be 3-manifolds. However, each data point that represents a particular discourse or story could be high-dimensional. In both discourse and story spaces, each data point will be an ordered set of moments, which can be considered three functions. A particular narration will be represented as a  set of all the data points that come from the same narration.  

To conceptually apprehend the 3-manifolds in high dimensions,  consider closed curves (i.e. 1-manifolds) in 3D. Unlike curves in 2D, the curves in 3D may not be planar and can form knots and links. The 3-manifolds in higher dimensions will be similar. In their local neighborhood, they will behave like Euclidean 3D,  but the global structure can be very complicated. We also expect that the collected data will not provide all possible areas in perceptual space since the storytellers always make intelligent assumptions about audience sensibilities and avoid the type of discourse and stories that are obviously not viable financially. 

\subsection{Contributions}

In this paper, we have six major contributions: 

\begin{enumerate}
\item We have introduced an approach to analyzing the effectiveness of stories. 
\item We have introduced the concept of moments as emotions evoked by characters and stories on the audience. 
\item We have classified moments into two main categories, Discourse and Story. 
\item For each of the two categories we have identified three types of universal moments. For Discourse, we have identified moments of: Concern, Endearment, and Justice. And for Story, we have identified moments of:  Curiosity, Surprise, and Clarity.
\item We have demonstrated that universal moments can be represented as functions and a visual story can be turned into a few sets of functions representing the collection of Discourse and Story type moments. The collection of moments can therefore be evaluated using function, statistical, and data analysis tools.
\item We have also demonstrated that universal moments of each Discourse and Story can be combined to only one lump-sum variable, allowing both Discourse and Story-type moments to be represented as a single-valued function, that can even be easier to analyze.  
\end{enumerate}

We have postulated that these perceptual discourse and story spaces are compact 3D Manifolds in high dimensions. Therefore, these spaces can serve a role similar to perceptive color spaces such as CIE-XYZ to organize and classify stories in a formal way. Such a structure can help to discover unexpected relationships between narrative structures. We can also compare and contrast the relative importance of discourse and story. It can also be easier to find niche movies that are attractive only for small and specialized markets. The opposite can also be correct. Niche markets for unusual stories can be identified by analyzing this space. 

%-------------------------------------------------------------------------
\section{Related Work}

Narratological analysis was started in the 1920s by Vladimir Propp \cite{Propp1973}, who developed a grammar covering a restricted corpus of Russian folktales. Propp's analysis decomposed a candidate story into an initial state comprising a small collection of characters (\emph{dramatis personae}) and a set of narrative functions over states. The application of a function to a state produces either the end state (the end of the story) or a new state. Propp showed that a small set of about 30 narrative functions plus a few constraints on function ordering could generate the whole chosen corpus of Russian folktales. Propp's analysis was used in some early story-telling programs in Artificial Intelligence (\cite{Meehan1977, Lebowitz1984, Lebowitz1985}, with limited results. Propp's theory was substantially refined in the 1960s (\cite{Barthes1966, Bremond1973}) when a distinct discipline called ``narratology'' emerged. 

Algirdas Julien Greimas introduced the concept of ``actant'' in place of Propp's characters and showed that a generic story could be analyzed in terms of the circulation, which is regulated by strict rules, of valuable objects among a very limited number of actants. Artificial Intelligence research in story-understanding and story-telling ignored post-Propp narratological research until very recently. Partly as a result of the work of Herman \cite{Herman2002} and Ryan \cite{Ryan2004}, computational approaches to narrative (\cite{Meister2005}) have gained a renewed impetus and the computational representation of standard narratological models is one of the explicit goals in the field (\cite{Scharfe2000, Peinado2004, whitman2009, Lakoff2010}).

Narrative theoreticians agree that there are at least two levels in any narration: Some events happen and these events are related in a certain way. Although there exist various terminologies used by different researchers, these two levels of a narration can be identified by two questions: (1) What is told and (2) How is it told? In the most widely used structuralist terminology, the answer to the ``what'' question is called a \textit{story} and the answer to the ``how'' question is called a \textit{discourse} \cite{chatman1980}.

% \begin{figure}[htbp]
% \centering
% \includegraphics[width=0.99\columnwidth]{images/framework}
% \caption{ Flowchart of the three layers and causality relationships among the layers according to Akleman et al. \cite{akleman2015theoretical}. Narration happens in the observation layer. It comes from the limited knowledge of actants (characters) and triggers further events through feedback to the physical layer. Narration stems from feedback mechanisms that trigger events based on the personalities of actants and observed narration. Visualization depends on both expression and observation layers. The expression layer provides information about the shapes and materials of characters. The observation layer provides information about cameras and shots. }  
% \label{fig:framework}
% \end{figure}

Since both story and discourse can be formulated as a series of cause-and-effect relationships,  the causal inference theory introduced by Judea Pearl \cite{pearl2000} gives a theoretical basis for representing stories as extended versions of directed acyclic graphs. These graphs can provide a representation of causal inferences used to describe narrative functions\cite{akleman2015theoretical}. Akleman et al. provide one such example of extended causal inference by defining three layers to provide precise answers to both ``what'' and ``how'' questions. 

The problem with causality-based approaches is that the interactions and expressions of the physical and emotional states of the characters can be extremely complicated. They can be represented by directed acyclic graphs where every vertex is a general graph. The number of elements in the graphs can also be very high. Therefore, many of the existing approaches are difficult to use for story analysis. In this work, we have developed a simple approach by looking at the problem from the audience's perspective. Instead of developing models for representing all possible stories, our goal is to model a limited set of reactions of the audience towards a given story.

\section{Discourse \& Story Moments}
In this section we will provide an overview of both Discourse and Story Moments, details and examples will be provided in a later section.

\subsection{Brief Overview to Universal Moments of Discourse}
\label{section:DiscourseMoments}

We have identified how the story is told can be formulated as an attachment to the individual characters in the story since the actions and emotions of characters define how the story is told. We have identified that there are mainly three types of evoked emotions towards a character. These are (1) moments of concern. (2) moments of endearment, and (3) moments of justice.   These three can be considered three linearly independent axes between $1$ and $-1$, in which one represents positive emotion and minus one represents negative emotion. Figure~\ref{figure/images/LadyBird_Marion_A0} provides a visual example of how a collection of Discourse moments for a specific character can be visualized along the film's timeline. 

The moments of concern can be conceptualized as an axis of pity \& envy \cite{ben1993envy,stevens1948envy}.  A positive number corresponds to some type of concern such as pity, $0$ corresponds to neutral, and a negative number corresponds to some type of negative concern such as jealousy or envy.  

The moments of endearment can be conceptualized as an axis of love \& hate \cite{hamlyn1978phenomena,balint1952love}. A positive number corresponds to some type of endearment such as love or like, $0$ corresponds to neutral, and a negative number corresponds to some type of negative endearment such as hate, or disgust. 

The moments of justice can be conceptualized as an axis of comeuppance \& getting away with an unflattering moment \cite{sugiyama2008interest, zimmerman1976proud}. A positive number corresponds to any type of served justice such as comeuppance, karma, or punished crime. On the other hand, a negative number corresponds to some type of withheld, unserved, or denied justice such as getting away with a crime or wrongdoing.

\subsection{Brief Overview to Universal Moments of Story}

Literary theory and psychology research strongly suggest that emotions evoked by the cause and effect relationships are also important to analyze the stories \cite{di2019curiosity,vogl2019surprise,vogl2020surprised,oy2023}. Discourse moments allow us to evaluate each character independently of each other. Cause and effect relationships, on the other hand, provide an additional dimension that comes from the structure of the plot resulting from the interaction between the characters. We, therefore, call these types of emotions Story moments. 

Based on existing literature, we have identified that there are mainly three types of evoked emotions towards a story. These are (1) moments of curiosity, (2) moments of clarity, and (3) moments of surprise \cite{di2019curiosity, vogl2019surprise, vogl2020surprised, oy2023}.  These three Story moments can also be considered three linearly independent axes between $1$ and $-1$, where one represents positive emotion and minus one represents negative emotion. 

The moments of curiosity can be conceptualized as an axis of curiosity \& apathy.  A positive number corresponds to some type of genuine interest such as curiosity and inquisitiveness and a negative number corresponds to some type of disinterest and indifference such as apathy.  

The moments of clarity can be conceptualized as an axis of clarity \& confusion.  A positive number corresponds to some sort of clear and coherent storytelling with simple, accurate, and structured organization. On the other hand, a negative number corresponds to confusing, disorientating, and distracting storytelling.  

The moments of surprise can be conceptualized as an axis of surprise \& predictable.  A positive number corresponds to an unpredictable event that evokes surprise, awe, and amazement. On the other hand, a negative number corresponds to an anticipated event that makes the audience bored since they can easily predict what can happen next.  

\section{Representations of Moments}

Our theoretical framework is based on a mathematical representation of a single moment. We observe that moments can be considered either a point or a vector. These two different representations may appear to be similar but the operations over them lead to two different types of results. In this paper, we mainly represent moments as vectors. Representation of moments as points is briefly discussed in the section~\ref{Sec/Discussion}.

\subsection{Moments as Vectors}

A moment vector is defined as a 3D vector
$\textbf{m} = (m_0, m_1, m_2) $
where $m_0, m_1, m_2 \in [-1,1]$ and $m_0, m_1, m_2$ are universal moments belonging to either Discourse or Story.
We assume that we can use any vector operation to analyze Discourse or Story moments. For instance, we can use the dot product to compute the similarities of two moments. We do not immediately see the use of some operations such as vector multiplication or rotation. On the other hand, they may still have some use. For instance, with $90^0$ degree rotations in $x$, $y$, and $z$ axis, one type of moment can be converted into another, which may have some application to identify similarities. 

\begin{figure}[htb!]
    \centering
        \begin{subfigure}[t]{0.49\textwidth}
        \includegraphics[width=1.0\textwidth]{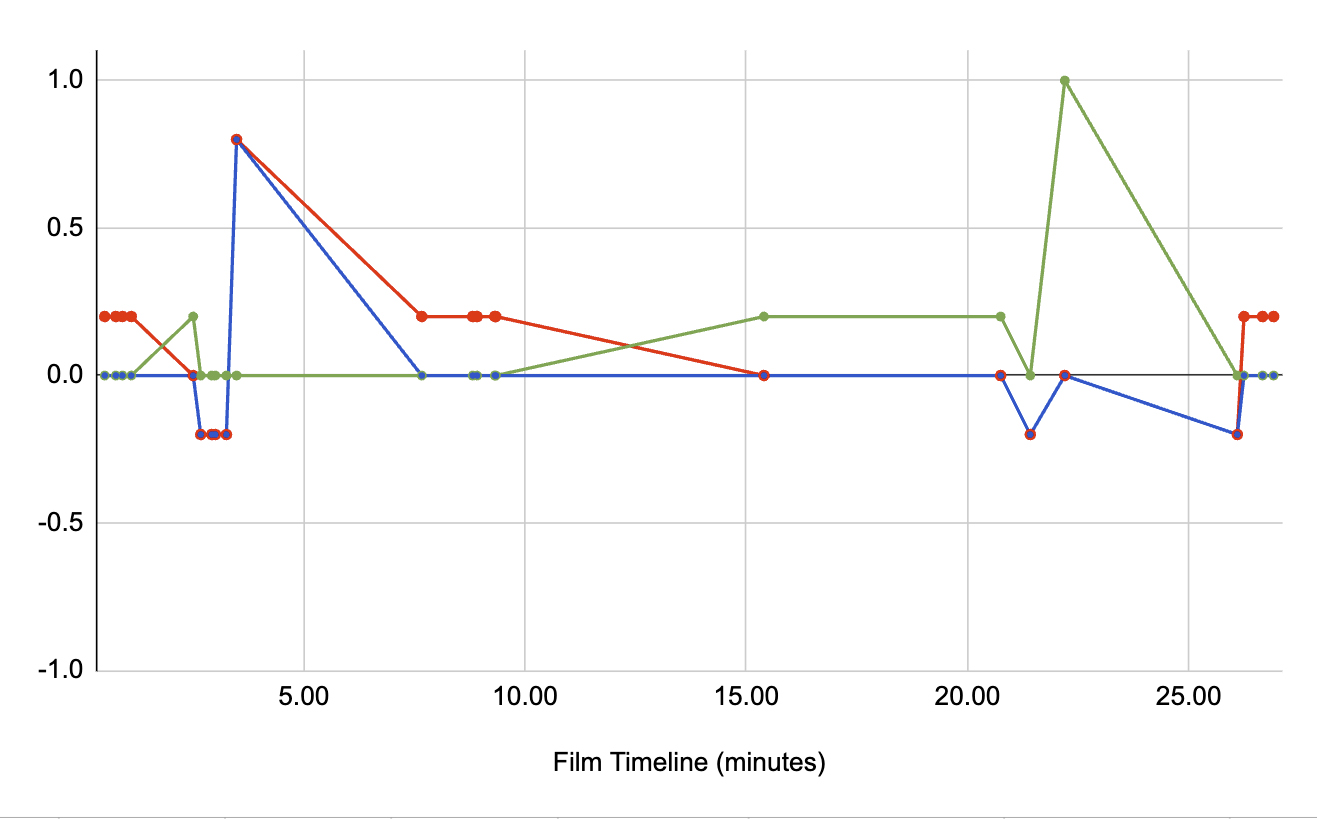}
\caption{This figure provides the visualization of Discourse universal moments as three continuous functions that linearly interpolate the set of given moments. The three types of universal moments $m_0, m_1$, and $m_2$ are drawn in Red, Green, and Blue respectively.   }
\label{figure/images/LadyBird_Marion_A0}
    \end{subfigure}
    \hfill
    \begin{subfigure}[t]{0.49\textwidth}
    \includegraphics[width=1.0\textwidth]{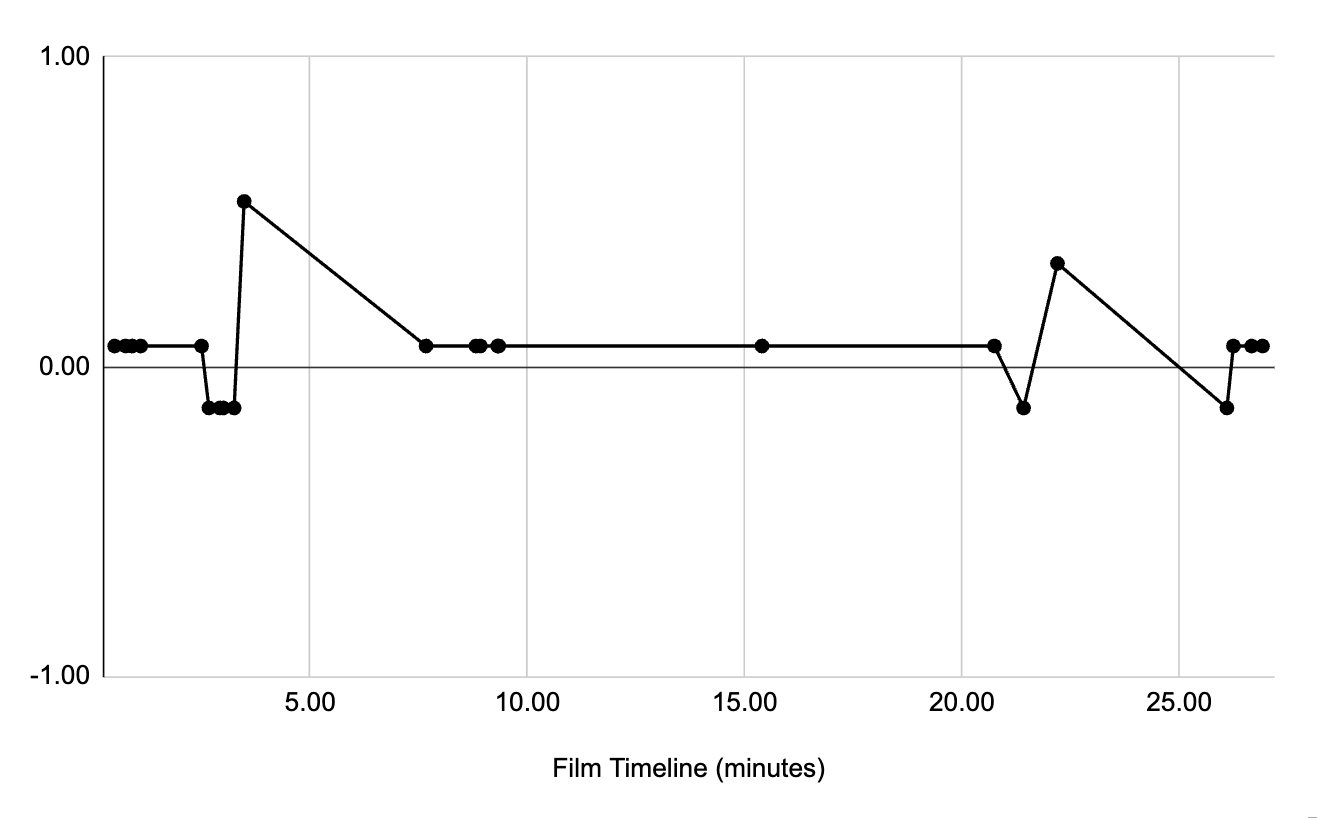}
\caption{This figure provides a visualization of Discourse moments as a single continuous function $\overline{f(t)}$ that is computed by combining the three functions in Figure~\ref{figure/images/LadyBird_Marion_A0}, using Equation~\ref{Equation/Barycentric}. 
}
\label{figure/images/LadyBird_Marion_B0}
    \end{subfigure}     
\caption{Visualization of a collection of Discourse moments. These two examples are based on the Discourse moments for the "Marion" character in the "Lady Bird" movie, which is widely analyzed in the literature \cite{stone2022lady, herdayanti2021psychological, musdalifa2022analysis}. Note in Figure~\ref{figure/images/LadyBird_Marion_B0} that moments fostering a sentiment of repulsion toward the character are quickly followed by moments fostering an attraction (see timeline at around minutes 3, 22, and 27).
}
\label{figure/images/LadyBird_Marion_Instant}
\end{figure}

Consider we have $N$ number of ordered moments, $ \textbf{m}_k=( m_{k,0}, m_{k,1}, m_{k,2})$ where $k=1,2,\ldots,N$, where $k$ corresponds to time $t_k$. We can visualize these moments as three continuous moment functions by linearly interpolating each moment as follows: 

\begin{eqnarray}
f_i(t) &=& m_{k,i} \frac{t_{k+1} -t}{t_{k+1} - t_k }  + m_{k+1,i} \frac{t-t_k}{t_{k+1} - t_k } \; \; \; \forall i \in \{0,1,2\}  \label{Equation/interpolate}\\
&\mbox{where}&\; \; t_k \; \; \mbox{corresponds to the time moment k occured.}  \nonumber \\
&\mbox{with}&\; \; t_k \in \Re^+ \; \; \mbox{and} \; \;  t_k < t_{k+1} \forall k \nonumber \\
\end{eqnarray}
The linear interpolation formula given in Equation~\ref{Equation/interpolate} is actually the equation of a first-degree non-uniform B-Spline function. Furthermore, 
these three functions form a parametric curve in 3D. 
If necessary, this can also be generalized to any non-uniform B-spline function to obtain a smoother approximation of the underlying data and the recursive version of this formula gives us B-spline curves in 3D. The only problem with this approach, it will be hard to understand the data since the time component will be missing. We instead draw them in 2D by providing time data explicitly and we assign different colors (red, green, and blue) to each function. An example of these functions is shown in Figure~\ref{figure/images/LadyBird_Marion_A0}. 

Since our basic goal is to identify the overall attraction and repulsion of the audience to the character or the story, this visualization can further be simplified into a single function by computing the Barycentric average of the three-moment functions as follows:

\begin{eqnarray}
\overline{f(t)} &=&   \sum_{i=0}^{2}  a_i f_i(t)  \nonumber \\
&\mbox{where}&\; \; 0 \leq  a_i \leq 1 \; \; \mbox{and} \; \; \sum_{i=0}^{2}  a_i =1 %\nonumber  
\label{Equation/Barycentric}
\end{eqnarray}

An example of $\overline{f(t)}$ is shown in Figure~\ref{figure/images/LadyBird_Marion_B0}. Note that this transformation significantly simplifies observing the overall level of attraction or repulsion that is being relayed by the story at a given point in time, for the given character.
The function $\overline{f(t)}$ is useful but it does not demonstrate the accumulation effect.  The degree of attraction or repulsion the audience experiences, at any given time, is based on the sum of the audience's experiences up until that given time. Therefore, we further realized that it is also useful to visualize the accumulation of moments.

\subsection{Accumulation of Moments}

We observe that the cumulative addition is a useful operation to grasp underlying moment data. If we add all the moments as vectors, we obtain the following formula for cumulative addition: 
\begin{eqnarray}
\textbf{M}_K =\left( M_{K,0}, M_{K,1}, M_{K,2} \right)  =  \left( \sum_{k=1}^{K}  m_{k,0}, \sum_{k=1}^{K}  m_{k,1}, \sum_{k=1}^{K} m_{k,2} \right) \label{Equation/0}
\end{eqnarray}
where $1 \leq K \leq N$. We again compute continuous functions by interpolating these values as follows: 

\begin{eqnarray}
F_i(t) &=& M_{K,i} \frac{t_{K+1} -t}{t_{K+1} - t_K }  + M_{K+1,i} \frac{t-t_K}{t_{K+1} - t_K } \forall i \in {0,1,2}  \label{Equation/2}\\
&\mbox{where}&\; \; t_K \; \; \mbox{corresponds the time of moment K occured.}  \nonumber \\
&\mbox{with}&\; \; t_K \in \Re^+ \; \; \mbox{and} \; \;  t_K < t_{K+1} \; \; \; \forall K \nonumber \\
\end{eqnarray}
Figure~\ref{figure/images/LadyBird_Marion_C0} provides an example of three accumulated-moment-functions based on moments given in Figure~\ref{figure/images/LadyBird_Marion_A0}. Note that this visualization provides a clearer view of the audience's attraction or repulsion for each type of universal moment. We can simplify these three functions further by computing their Barycentric average as follows:

\begin{figure}[htb!]
    \centering
        \begin{subfigure}[t]{0.49\textwidth}
        \includegraphics[width=1.0\textwidth]{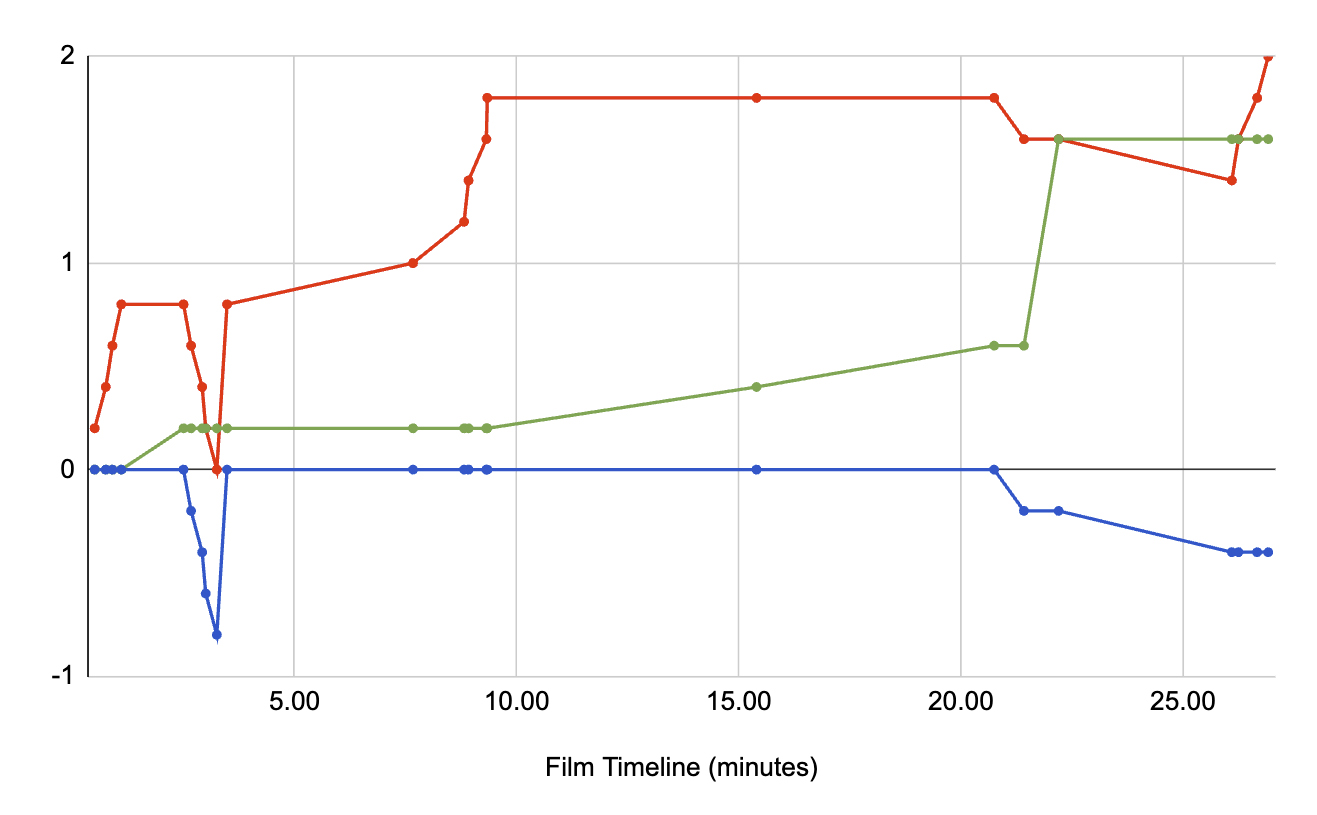}
\caption{This figure provides the visualization of three accumulated universal moments as three continuous functions that linearly interpolate a set of accumulated moments originating from data in Figure~\ref{figure/images/LadyBird_Marion_A0}. }
\label{figure/images/LadyBird_Marion_C0}
    \end{subfigure}
    \hfill
    \begin{subfigure}[t]{0.49\textwidth}
    \includegraphics[width=1.0\textwidth]{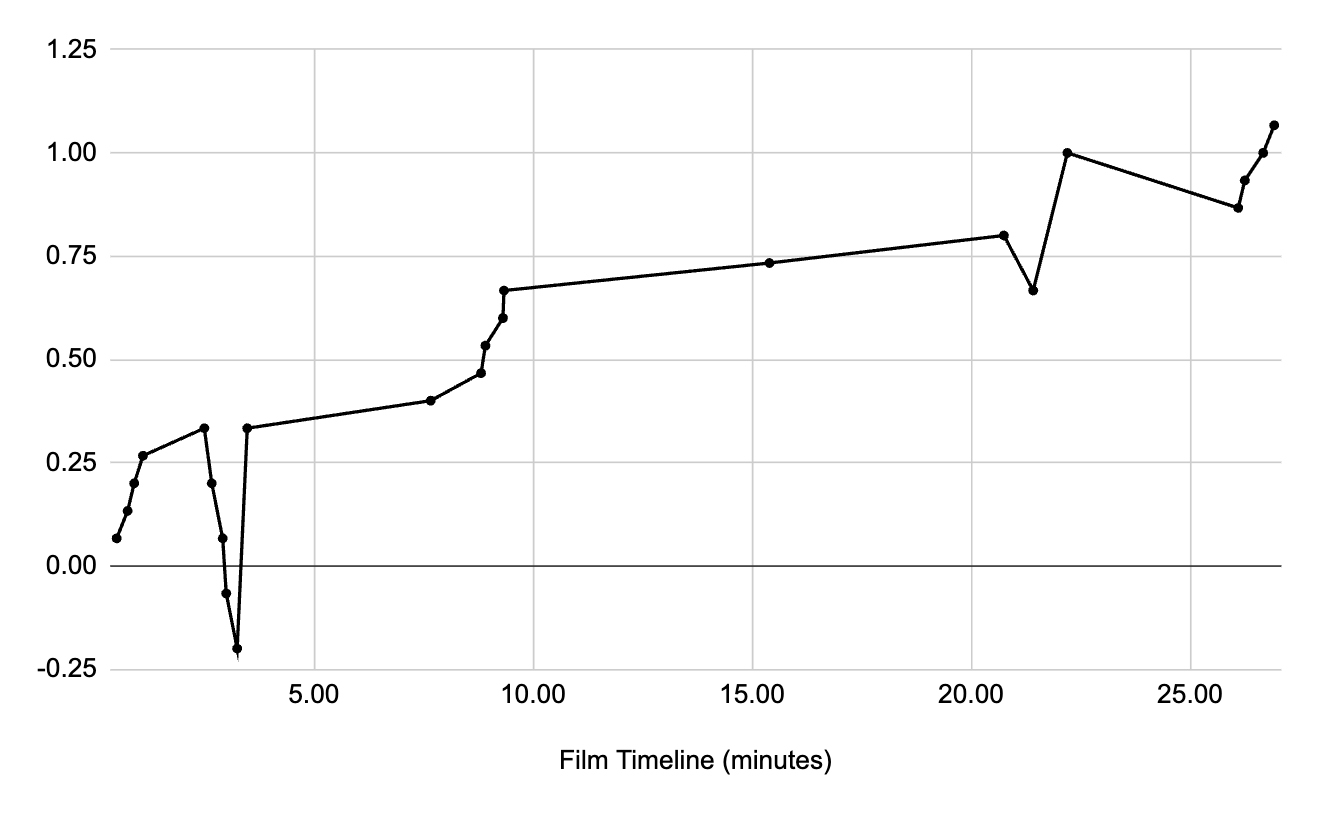}
\caption{This figure provides an example of $\overline{F(t)}$ for the data in Figure~\ref{figure/images/LadyBird_Marion_A0}. Notice the ease in comprehending the overall level of attraction or repulsion toward the character at any given time.}
\label{figure/images/LadyBird_Marion_D0}
    \end{subfigure}     
\caption{Visualization of Accumulated Discourse moments for the "Marion" character in the "Lady Bird" movie. The three functions, $F_0(t), F_1(t),$ and $F_2(t)$, that correspond to three accumulated moments $M_0, M_1$, and $M_2$ are drawn in Red, Green, and Blue respectively. Notice the ease in deciphering which type of universal moments (Red, Green, or Blue) contribute most to the attraction and repulsion at any given time.  }
\label{figure/images/LadyBird_Marion_Accumulation}
\end{figure}

\begin{eqnarray}
\overline{F(t)} &=&   \sum_{i=0}^{2}  a_i F_i(t)  \nonumber \\
&\mbox{where}&\; \; 0 \leq  a_i \leq 1 \; \; \mbox{and} \; \; \sum_{i=0}^{2}  a_i =1 \nonumber  \label{Equation/Barycentric2}
\end{eqnarray}

An example of $\overline{F(t)}$ is shown in Figure~\ref{figure/images/LadyBird_Marion_D0}. The function $\overline{F(t)}$ is particularly useful since it demonstrates the accumulation effect of attraction and repulsion in a single function.

In these examples, we chose Greta Gerwig's \emph{Lady Bird} movie since there was significant interest in that movie in the literature \cite{weiss2009teaching}. Since moments can be negative, the accumulated moment functions can have both increasing and decreasing portions. Increasing portions suggest that the audience's positive attachment to the character in that duration is increasing. The decreasing function suggests that the audience's negative attachment to the character in that duration is increasing. The function-based visualization, therefore, is useful to see the overall change. %On the other hand, color charts, detailed later in Section \ref{Color Charts Section} are useful to see the type of moments in a given time. 

\section{Audience-Character Emotional Attachment and Discourse Moments}

In this section, we provide a more detailed view of the Discourse moments.  The main premise of the Discourse moments is to provide a measure of "Audience-Character Emotional Attachment", which we defined as the degree to which the audience is attracted, indifferent, or repulsed by a given character.  

Inspired to better understand the reasons why an audience might become emotionally attached to a given character we set out to watch over 50 studio films meticulously looking for any moment in the story we felt contributed positively or negatively to the audience-character emotional attachment.
We also reviewed psychology and humanities literature to identify the types of audience-character emotional attachment.
From this investigation we identified that there are three main types of universal storytelling moments, that contribute significantly to establishing, maintaining, strengthening, or deteriorating the audience’s emotional attachment to a character. We found that these three types of moments, associated with a given character, cumulatively build on each other to influence and evolve the emotional attachment of the audience to the given character.  

Each of the three types of universal moments has been given a name for ease of reference along with a definition to explain the particular type of moment.  Additionally, the three types of moments listed below are not specific to any one feature film genre, rather they are universal moments that apply to all genres. These three universal moments, as presented earlier,  are concern, endearment, and justice. 

\subsection{Moments of Concern}

Concern as discussed earlier can be viewed as an axis of pity \& envy. These are moments in the story that could foster a range of feelings from sympathetic pity to negative envy or jealousy from the audience to the character. 

An example of a moment of positive concern (pity) in \emph{Night At The Museum} \cite{levy2006} is the moment when we see Ben Stiller’s character receive a parking ticket and wheel lock on his car. In \emph{Back To The Future} \cite{zemeckis1985}, another example of a moment of positive concern in the type of pity is the moment when Marty McFly’s high school principal gets in Marty’s face telling him sternly “I noticed your band is on the roster for the dance auditions after school today. Why even bother McFly?! You don't have a chance, you're too much like your Old Man. No McFly has ever amounted to anything in the \emph{history} of Hill Valley!” 

Moments of positive concern also include moments that convey misfortunes that can noticeably be greater in magnitude than the smaller moments of pity mentioned above.  These include moments such as when the story conveys that the character is, or has in the past, experienced something capable of causing substantial emotional pain, hurt, or trauma to the character. This scale of misfortune can generate a large dose of pity for a character.  Additionally, for characters that are behaving in an off-putting way, this larger scale of misfortune, if it appears to be the underlying source or origin of the off-putting behavior, can help generate compassion toward the character. This compassion will help \emph{diffuse} the emotionally repelling feelings that may have otherwise formed in the audience toward the character due to their off-putting behavior. 

In \emph{Finding Nemo}\cite{stanton2003} the first four and a half minutes of the film are dedicated to illustrating a large misfortune for Marlin when all but one of his family members are killed by a predator fish. Therefore, later when Marlin is exhibiting his overprotective, overzealous parenting approach we have some understanding and compassion and are not outright repelled by his off-putting behavior. 

In \emph{Lady Bird} \cite{gerwig2017} we see this type of moment for (Lady Bird’s mother) Marion come and go in the span of a few seconds when Marion tells Lady Bird “My mother was an abusive alcoholic.” We can imagine that it was likely emotionally painful or traumatic as a child to be raised by an abusive alcoholic mother. We also learn, by way of this moment, that Marion did not have a suitable role model for parenting. Like Marlin from \emph{Finding Nemo}, this moment gives us some understanding and compassion for her off-putting parenting style.

\subsection{Endearing Moments} 

These are moments in the story that could foster positive feelings such as attraction, endearment, respect, love, affection, fondness, admiration, reverence, or equivalent from the audience to the character. On the negative side, they can foster negative feelings such as hate and disgust based on an unflattering moment. 

In \emph{Iron Man} it is an Endearing moment when Tony Stark, who is riding in a military vehicle with several quiet soldiers, breaks the awkward silence by telling funny jokes that makes everyone in the vehicle smile and laugh \cite{favreau2008}. In \emph{Die Hard} it is an Endearing moment when John McClane chooses to sit up front in the passenger seat of the luxurious limousine that has picked him up from the airport \cite{mctiernan1988}.

The negative endearing moments in the story happen when the character says, does, or has done something that could be considered within the realm of unflattering, unbecoming, unattractive, unappealing, flawed, negative, unscrupulous, or equivalent. Unlike the positive moments that foster an emotional attachment toward a character, such negative moments have the potential to emotionally repel the audience from the character by evoking feelings such as hate or disgust. That said, \emph{Unflattering} moments are required for characters undergoing an inner transformation since it is through this particular type of moment that we experience the character exhibiting their flawed behavior. 

In \emph{Psycho} it is an \emph{unflattering} (or negative endearment) moment when Marion steals an envelope of cash from her workplace \cite{hitchcock1960}. And in \emph{A Bug's Life} it is an \emph{Unflattering} moment when Flik chooses to back down and away from Hopper who, in his anger, lifts Flik’s young friend Dot high off the ground by tightly gripping her head \cite{lasseter1998}. Dot just a few feet away from Flik, looks scared and winces, Flik chooses not to help her.

\subsection{Moments of Justice} 

The moments of justice can be considered an axis of \emph{Comeuppance} \& getting away with an unflattering moment. 
The comeuppance is a type of moment in the story when the character who has previously acted with negative behavior is the recipient of some negative event or circumstance that could be considered deserved criticism, payback, punishment, or karma for their negative behavior. We observe that instances of \emph{Comeuppance} diffuse and neutralize negative sentiments the audience may begin to have toward a character who is exhibiting negative behaviors. On the other hand, if a character continues getting away with an unflattering event, the negative attitude towards the character increases. We, therefore, realized that justice should be considered a separate dimension. 

In \emph{Toy Story} it is a moment of \emph{Comeuppance} when Woody in a jealous outburst, kicks a plastic Checkers piece which then ricochets off a nearby wall hitting him in the face \cite{lasseter1995}. And in \emph{Up} it is a moment of \emph{Comeuppance} when Carl is sentenced by the court to live at a retirement home after hitting a construction worker on the head with his cane.

\emph{Comeuppance} moments, like the other moments, can have varying scales of magnitude. For example, a much larger moment of \emph{Comeuppance} can be found in \emph{The Godfather Part II} for Michael Corleone when his wife Kay (who despises Michael's lawless lifestyle) reveals to Michael that she didn't have a miscarriage, she had an abortion because she didn't want their unborn son to grow up to be like Michael. 

\subsection{Recording and Visualization of Discourse Moments for a Single Character}
\label{Recording Discourse Moments} 
\begin{figure}[htb!]
    \centering
        \begin{subfigure}[t]{0.49\textwidth}
        \includegraphics[width=1.0\textwidth]{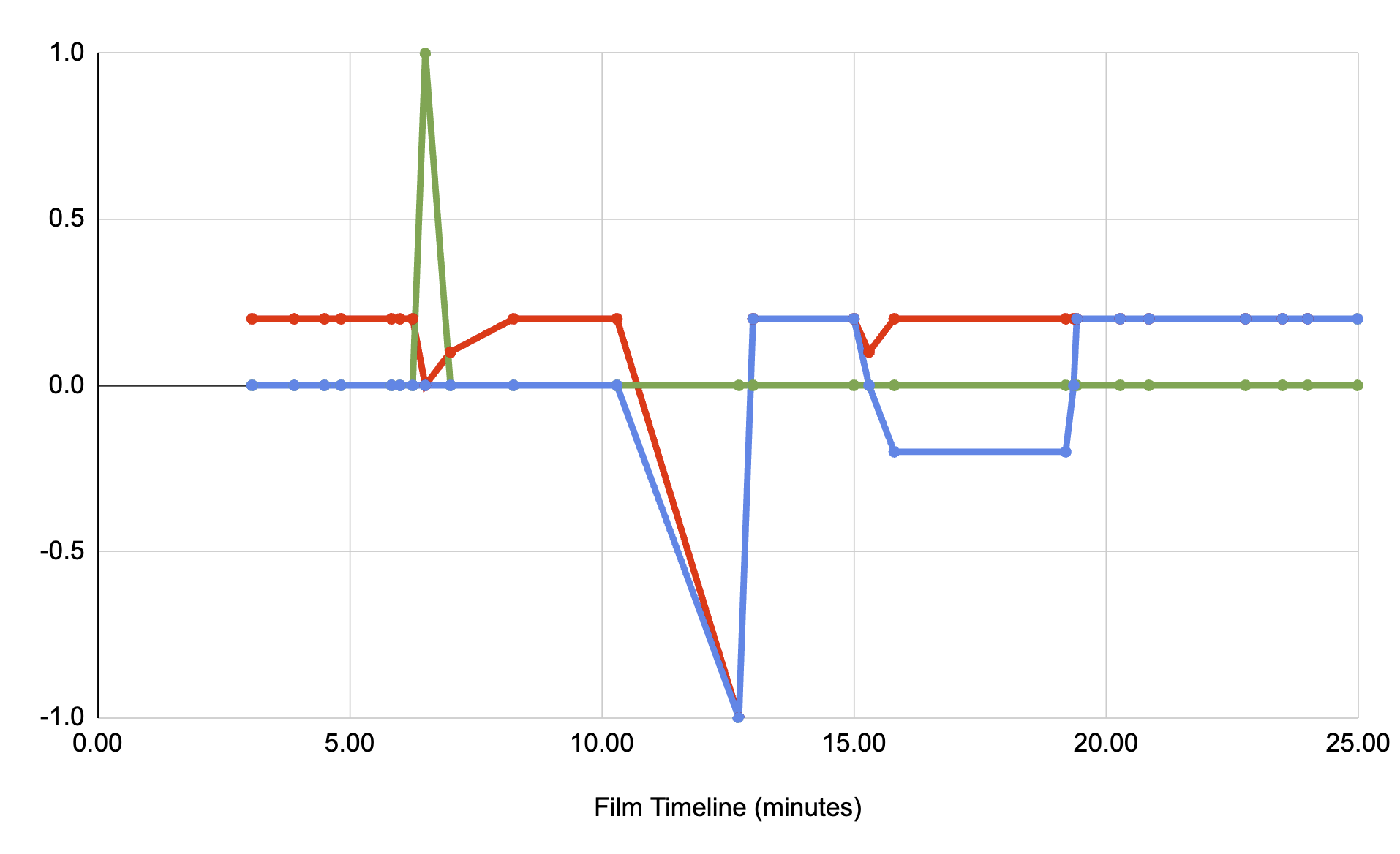}
\caption{Marion's Discourse moments as they are relayed to the audience. }
\label{figure/images/Psycho_Marion_A0}
    \end{subfigure}
    \hfill
    \begin{subfigure}[t]{0.49\textwidth}
    \includegraphics[width=1.0\textwidth]{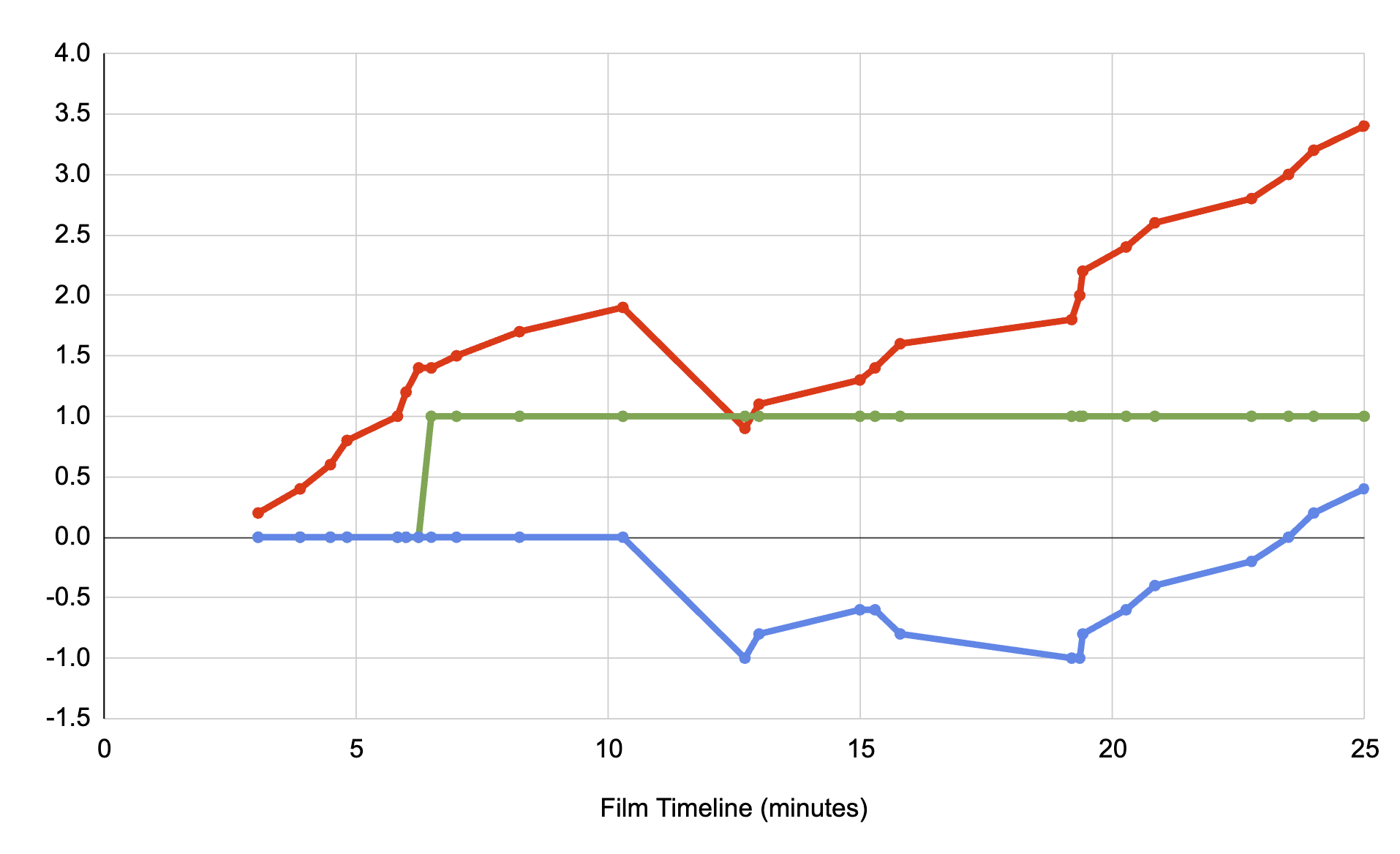}
\caption{Accumulation of Marion's Discourse moments.}
\label{figure/images/Psycho_Marion_C0}
    \end{subfigure}  
\caption{These two sets of functions provide the Discourse moments relayed by the story at a given time (Figure~\ref{figure/images/Psycho_Marion_A0}) and the accumulation of those moments (Figure~\ref{figure/images/Psycho_Marion_C0}) for the character Marion in \emph{Psycho}. The three major evoked emotions toward a character are illustrated as (Red) Endearment, (Green) Concern, and (Blue) Justice. This particular set of functions demonstrates how  Alfred Hitchcock controlled the audience's emotional attachment to Marion before she was killed, which would eventually be an ultimate comeuppance for her crime. }
%(Reviewers, please see detailed analysis in additional documents.)  }
 \label{figure/images/Psycho_Marion}
 \end{figure}

Since, these three main types of evoked emotions can be considered three linearly independent axes between $1$ and $-1$, where one represents positive emotion, zero represents neutral emotion, and minus one represents negative emotion, we are able to associate a numeric value with every instance of the three universal storytelling moments to indicate the perceived magnitude of that moment.  For entering these numbers, there is no need to be precise. However, their relative importance needs to be captured.  For instance, Ben Stiller's character in \emph{Night At The Museum} receiving a parking ticket may be given a smaller positive value such as 0.1 or 0.2 since this is not a major concern whereas the death of Marlin's family in \emph{Finding Nemo} be may given a much larger value such as 0.9 or 1.0. See Figure \ref{figure/images/Psycho_Marion} for an example of Discourse moments (accumulated and not) for the character Marion in \emph{Psycho} relayed during the first 25 minutes of the film, which includes events \emph{before} she stops for the night at the Bates Motel. 

\subsection{Analyzing the Discourse Moments with Multiple Characters}

The three types of universal Discourse moments are defined relative to only a single character. For example, imagine two characters: Jill and Bob. At 10 minutes into the film, an intoxicated Jill has a petty argument with Bob, then she punches him in the face. This moment in the film could be an unflattering (negative endearment) moment for Jill, while at the same time, being a moment of pity (positive concern) for Bob. 

If the analysis is dealing only with Jill then this moment would be recorded as both an \emph{Unflattering} (negative endearment) and a negative justice moment for Jill. Note that because Jill's negative behavior is not punished, this unflattering moment also spawns a negative justice moment. That negativeness with justice can stay and linger until a comeuppance moment comes and the audience feels that justice is served. Whereas, if the analysis is dealing only with Bob, then this moment would be recorded as a moment of positive concern (pity) for Bob. 

Note that recorded analysis of multiple characters also provides information between the characters, however, that information is not well defined since we do not record what transpired. Instead, we only record the emotions evoked by the characters. Therefore, there is also a need to record emotions evoked by cause-and-effect relationships independent of characters. In other words, Story moments, are described in section \ref{Story Moments}. 

\subsection{Comparison of Audience Attachment of Characters}

% Figure~\ref{varietyOfIntegrals} shows characters from a variety of movies. A positive slope on the curve indicates the possibility of a growing positive emotional attachment to a character, and a negative slope indicates the possibility of a diminishing emotional attachment. 

\begin{figure}[htb!]
\centering
         \includegraphics[width=0.9\textwidth]{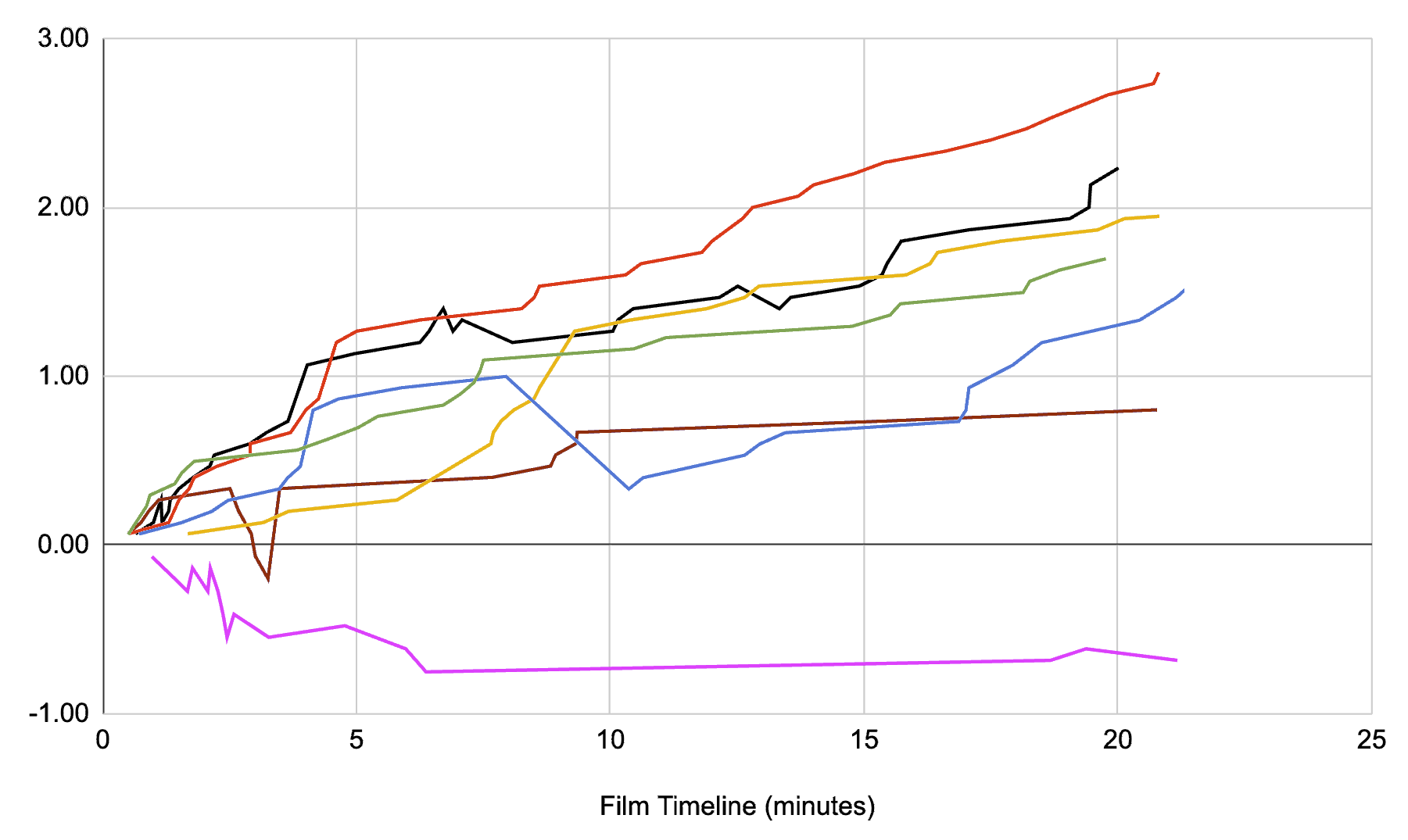}
\caption{This collection of accumulated moments shows accumulated attraction to the characters Billy Elliot, Tony Stark (Ironman), Luke Skywalker and Han Solo (Star Wars: A New Hope), Marion (Psycho), Marion (Lady Bird), and Valerian (Valerian and the City of Thousand Planets). The only negative accumulation is Valerian.  }
\label{figure/images/varietyOfIntegrals}
\end{figure}

Figure \ref{figure/images/varietyOfIntegrals} features the accumulated Discourse moments for eight characters in seven movies, each represented by a single curve showing the overall attraction. Curves have been assigned a random color to distinguish one from another. Higher values indicate an attraction toward the character and negative values indicate a repulsion toward the character. To facilitate comparisons the characters' original curves have been translated in the shared timeline such that their first moment occurs at minute 1.5. Note that the character Valerian from the film \emph{Valerian and The City of a Thousand Planets} resides completely in the negative territory.  

\begin{figure}[htb!]
\centering
         \includegraphics[width=0.8\textwidth]{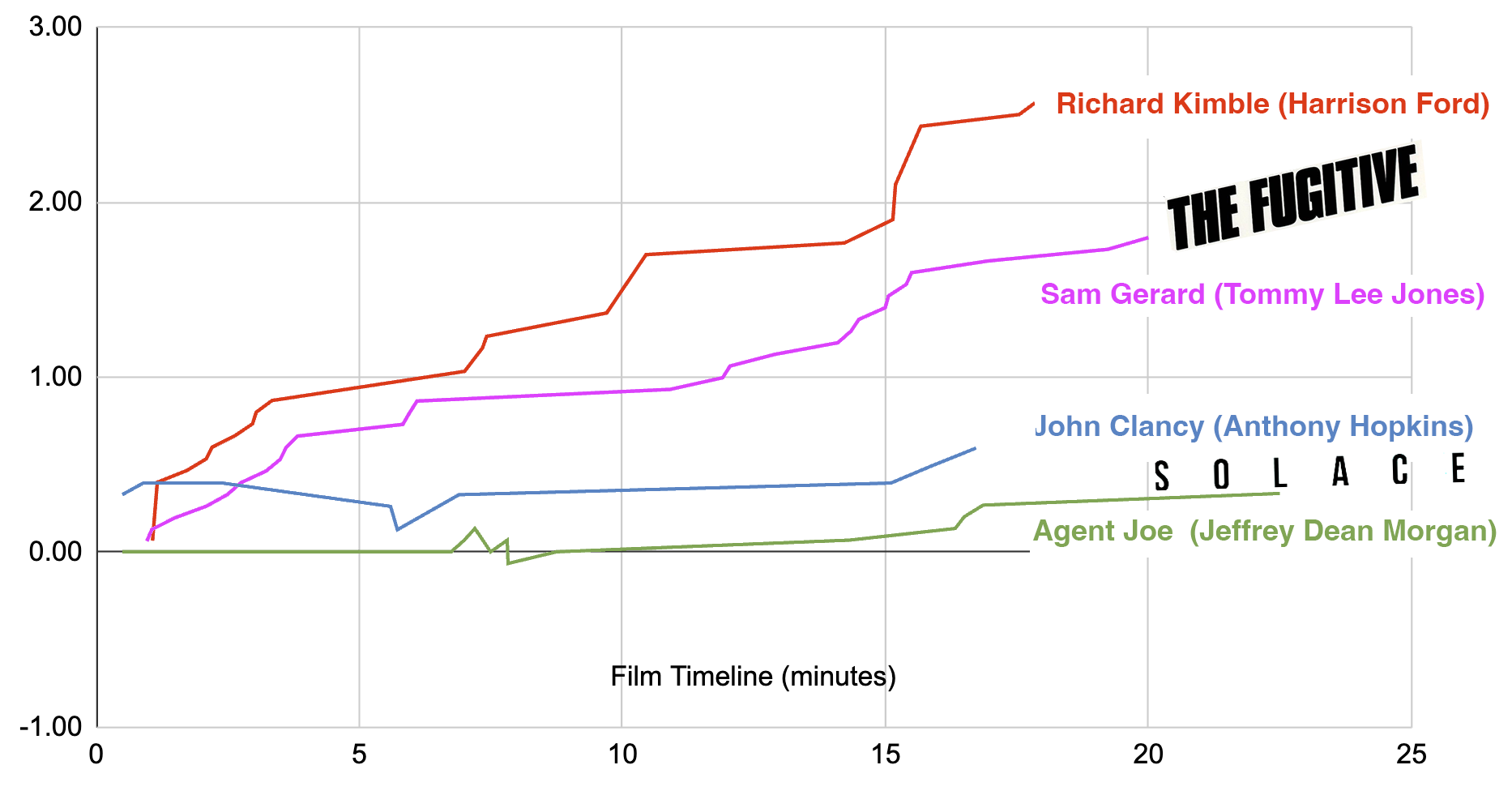}
\caption{This is a comparison of two main characters of two movies: The Fugitive and Solace. According to the Rotten Tomatoes page, The Fugitive's Tomatometer is 96\% and Audience Score is 89\%. On the other hand, Solace's Tomatometer is 24\% and Audience Score is 44\%. Our functions provide an explanation for these scores.  }
\label{figure/images/Fugitive_vs_Solace}
\end{figure}

Figure \ref{figure/images/Fugitive_vs_Solace} features the accumulated moments for four characters in two movies of the same genre (Thriller/Mystery), represented by their single overall attraction curves. Once again, curves have been translated to start at minute 1.5 in the shared timeline and have been assigned a random color to distinguish one from another.

\begin{figure}[htb!]
\centering
    \begin{subfigure}[t]{0.40\textwidth}
        \includegraphics[width=1.0\textwidth]{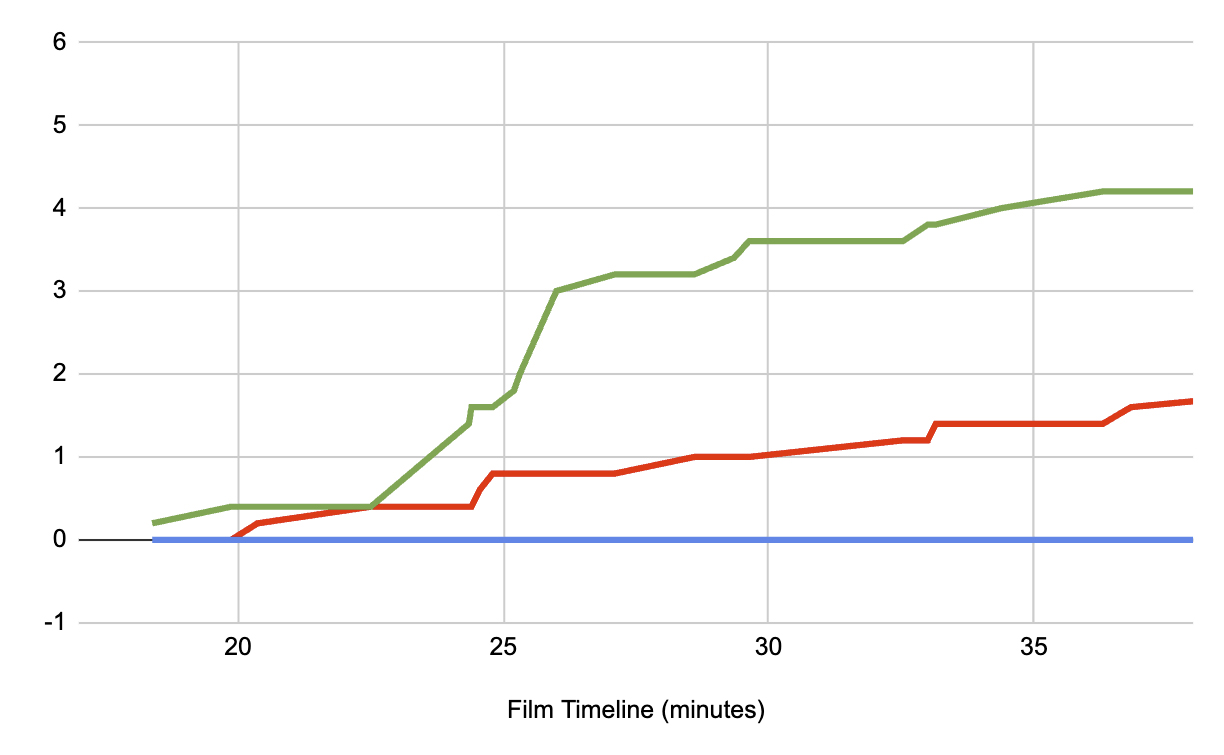}
        \caption{Accumulated Moments of Luke Skywalker}
        \label{figure/images/StarWars_Luke_C0}
    \end{subfigure}
    \hfill
    \begin{subfigure}[t]{0.58\textwidth}
        \includegraphics[width=1.00\textwidth]{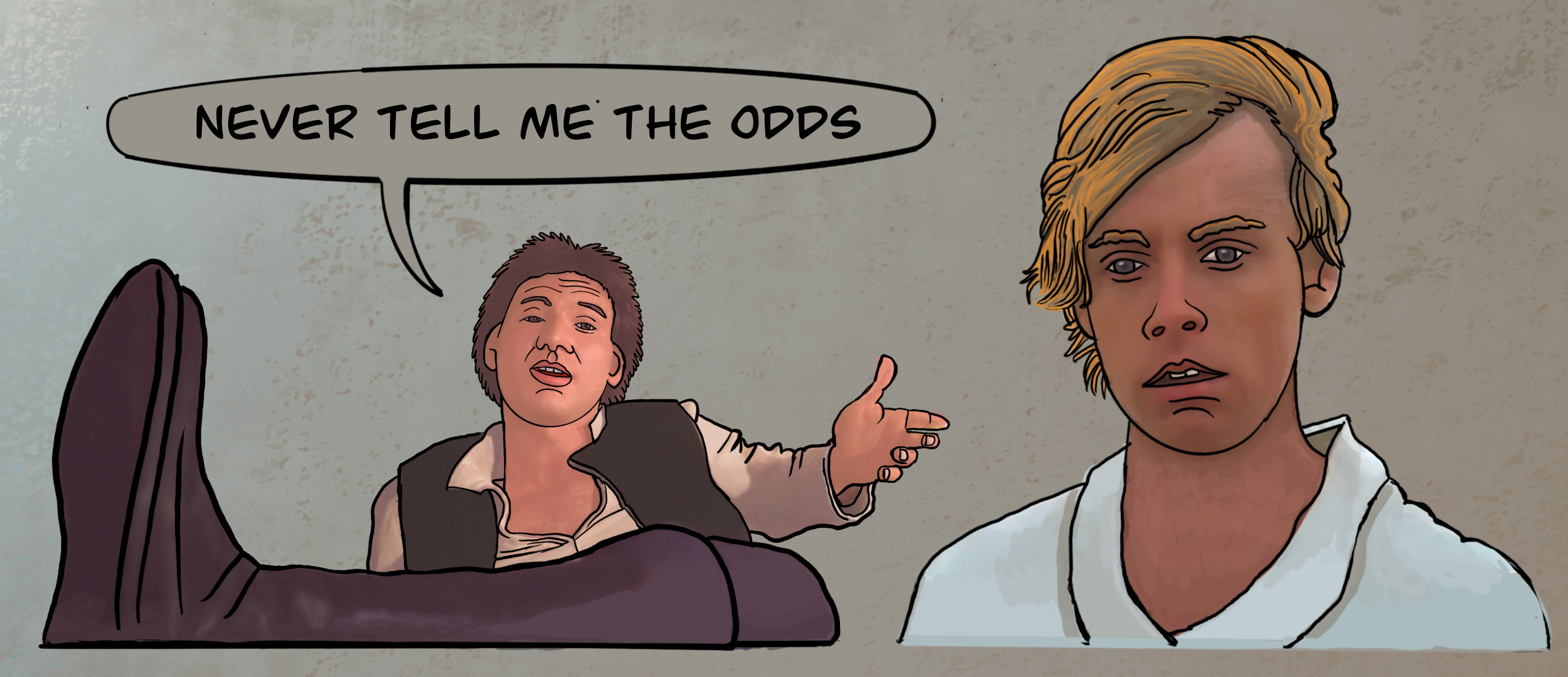}  
        \caption{A Visual Comparison of Luke Skywalker and Han Solo Characters}
        \label{figure/personalities0}
    \end{subfigure}
        \hfill
    \begin{subfigure}[t]{0.58\textwidth}
        \includegraphics[width=1.00\textwidth]{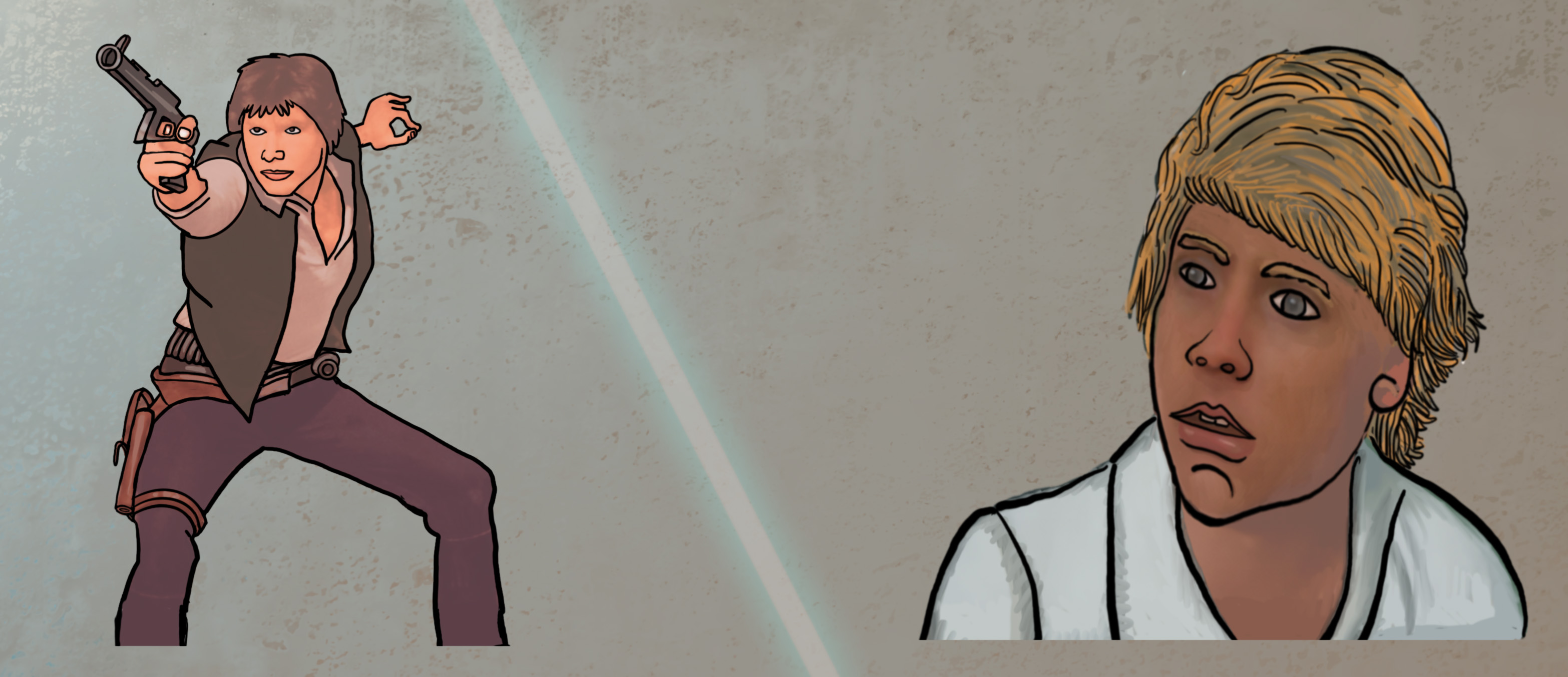}  
        \caption{Another Visual Comparison of the two Characters}
        \label{figure/personalities1}
    \end{subfigure}
        \hfill
    \begin{subfigure}[t]{0.40\textwidth}
        \includegraphics[width=1.0\textwidth]{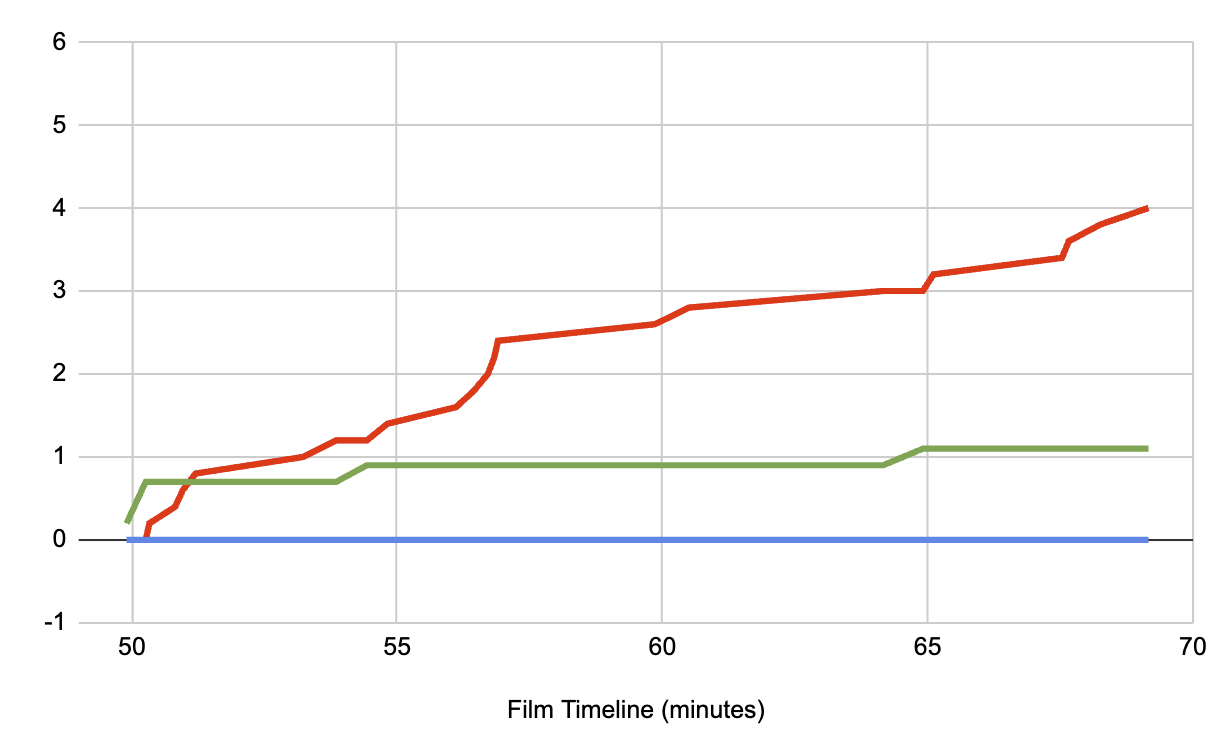} 
        \caption{Han Solo's accumulated Discourse Moments}
        \label{figure/images/StarWars_Luke_C1}
    \end{subfigure}
\caption{These set of functions illustrate the accumulated moments of Discourse for both Luke Skywalker and Han Solo in \emph{Star Wars: A New Hope}. In these graphs, Red is Endearment, Green is Concern, and Blue is Justice. Notice that both characters begin with moments of positive Concern (Pity) that are quickly followed by moments of positive Endearment (Love). Also, notice that in this time period, Luke's attachment heavily relies on moments of positive Concern, whereas Han Solo's attachment is based mostly upon moments of positive Endearment. This observation can be confirmed with visual evaluation of the two characters shown in \ref{figure/personalities0} and \ref{figure/personalities1}. The illustrations are done by one of the authors.}
\label{figure/images/StarWars_Luke_Han}
\end{figure}

Figure \ref{figure/images/StarWars_Luke_Han} features the accumulated  moments for the characters Luke Skywalker and Han Solo as seen in \emph{Star Wars: A New Hope} for the first twenty minutes of each character.

%\begin{figure}[ht!]
%\includegraphics[width=0.5\textwidth]{images/StarWars_Luke_C0}  
%\includegraphics[width=0.5\textwidth]{images/StarWars_HanSolo_C0}      
%\caption{These two sets of functions provide the Discourse moments relayed by the story at a given time (top) and the accumulation of those relayed moments (bottom) for the character Marion in \emph{Psycho}. The three major evoked emotions toward a character are illustrated as (Red) Endearment, (Green) Concern, and (Blue) Justice. This particular set of functions demonstrates how  Alfred Hitchcock controlled the audience's emotional attachment to Marion before she was killed, which would eventually be an ultimate comeuppance for her crime. (Reviewers, please see detailed analysis in additional documents.)  }
%\label{figure/images/StarWars_Luke_Han}
%\end{figure}
 
\section{ Audience-Story Emotional Attachment and Story Moments}
\label{Story Moments}
The main premise of the story moments is to provide a measure of "Audience-Story Emotional Attachment", which we defined as the degree to which the audience is attracted, indifferent, or repulsed by the story.

Inspired to better understand the reasons why an audience might become emotionally attached to a story we set out to watch over 50 studio films meticulously looking for any moment in the story we felt contributed positively or negatively to the audience's emotional attachment.
We also reviewed psychology and humanities literature to identify the types of audience-story emotional attachment. 
From this investigation we identified that there are three main types of universal storytelling moments, which contribute significantly to establishing, maintaining, strengthening, or deteriorating the audience’s emotional attachment to the story. 

Each of the three types of universal moments has been given a name for ease of reference along with a definition to explain the particular type of moment.  Additionally, the three types of moments listed below are not specific to any one feature film genre, rather they are universal moments that apply to all genres. These three universal moments, as presented earlier,  are surprise, curiosity, and clarity. We will evaluate each universal moment in a subsection. 

\subsection{Moments of Surprise}

The moments of surprise can be best conceptualized as an axis of surprise \& predictable.  Surprise can range from a smaller unpredictable event that is unexpected and intriguing to something relatively larger such as an unpredictable event that evokes surprise, awe, and amazement. On the other hand, predictable (or negative surprise)  corresponds to an anticipated event that makes the audience bored since they can easily predict what can happen next.

In \emph{Toy Story} there is a moment with a relatively small amount of \emph{Surprise} during Andy's birthday party when the Army soldiers give the all-clear that Andy is done opening presents. Woody and the other toys begin to feel safe and relaxed. But then! Andy's mom says "Wait a minute.  Oooh, what do we have here?!" as she reveals the actual last present, a Buzz Lightyear doll. In \emph{Star Wars: A New Hope} there is a relatively larger moment of \emph{Surprise} when Han Solo returns unexpectedly at the climax of Luke's X-Wing attack on the Death Star to lend Luke a helping hand. Prior, the story led the audience to believe that Han is on his to go pay-off some old debts and has zero interest in participating in the Death Star battle.

\subsection{Moments of Curiosity}
Curiosity as discussed earlier can be viewed as an axis of curiosity \& apathy. These are moments in the story that could foster a range of feelings from an intense eagerness to complete disinterest to learn what will be told next in the story. 

In \emph{Star Wars: A New Hope} it is a moment of curiosity when Luke, Leia, Han, and Chewie become trapped in a large trash compactor and suddenly the compactor walls begin to close in. In \emph{Finding Nemo} it is a moment of curiosity when Marlin's son, Nemo, is taken away in a boat after being captured by a scuba diver.

In the movie \emph{Valerian and the City of a Thousand Planets} it is a moment of apathy when Valerian reveals to Laureline that he knows where to go because the Princess has been giving him directions through visions in his head. Valerian is being given a solution that comes out of the blue, it comes across like Deus Ex Machina fostering some disinterest in his plight. 

\subsection{Moments of Clarity}

These are moments in the story that could foster positive feelings such as clarity and comprehension from the audience to the story. On the negative side, they can foster negative feelings such as dissatisfying confusion or bewilderment about some aspect of the story.

In \emph{Psycho} it is a moment of clarity when we learn what Marion's objective is when she drives her car with her packed suitcase and envelope of stolen cash. Through Marion's imagination, we hear her boyfriend Sam say "Marion? What in the world? What are you doing up here? Of course, I'm glad to see you, I always am..." In this moment we get clarity that Marion is embarking on a surprise visit to her boyfriend's house.

In \emph{Solace} it is a \emph{confusing} (or negative clarity) moment when FBI agents Joe, Katherine, and John, guns drawn, suddenly stop hunting for a serial killer who is on the loose in an apartment they are investigating. The trio, searching for a killer, see a dead body in the bathtub, then put their guns in their holsters and investigate the body. Why did they stop looking for the serial killer who is in the apartment with them?

\subsubsection{Simplifying Moments of Clarity}
\label{Simplifying Moments of Clarity}
Although it is completely valid to record both moments of Clarity and Confusion (negative clarity), we postulate that an accurate measurement of clarity and comprehension can be obtained by solely recording and analyzing moments of Confusion. In this approach, an assessment of clarity evoked by the story is simply based on the frequency and magnitude of moments of Confusion. The advantage of this approach is that data collection for Moments of Clarity is simplified while retaining its worth.

\subsection{Recording and Visualization of Essential Moments for Story}
 
\begin{figure}[htb!]
\centering
 \includegraphics[width=0.8\textwidth]{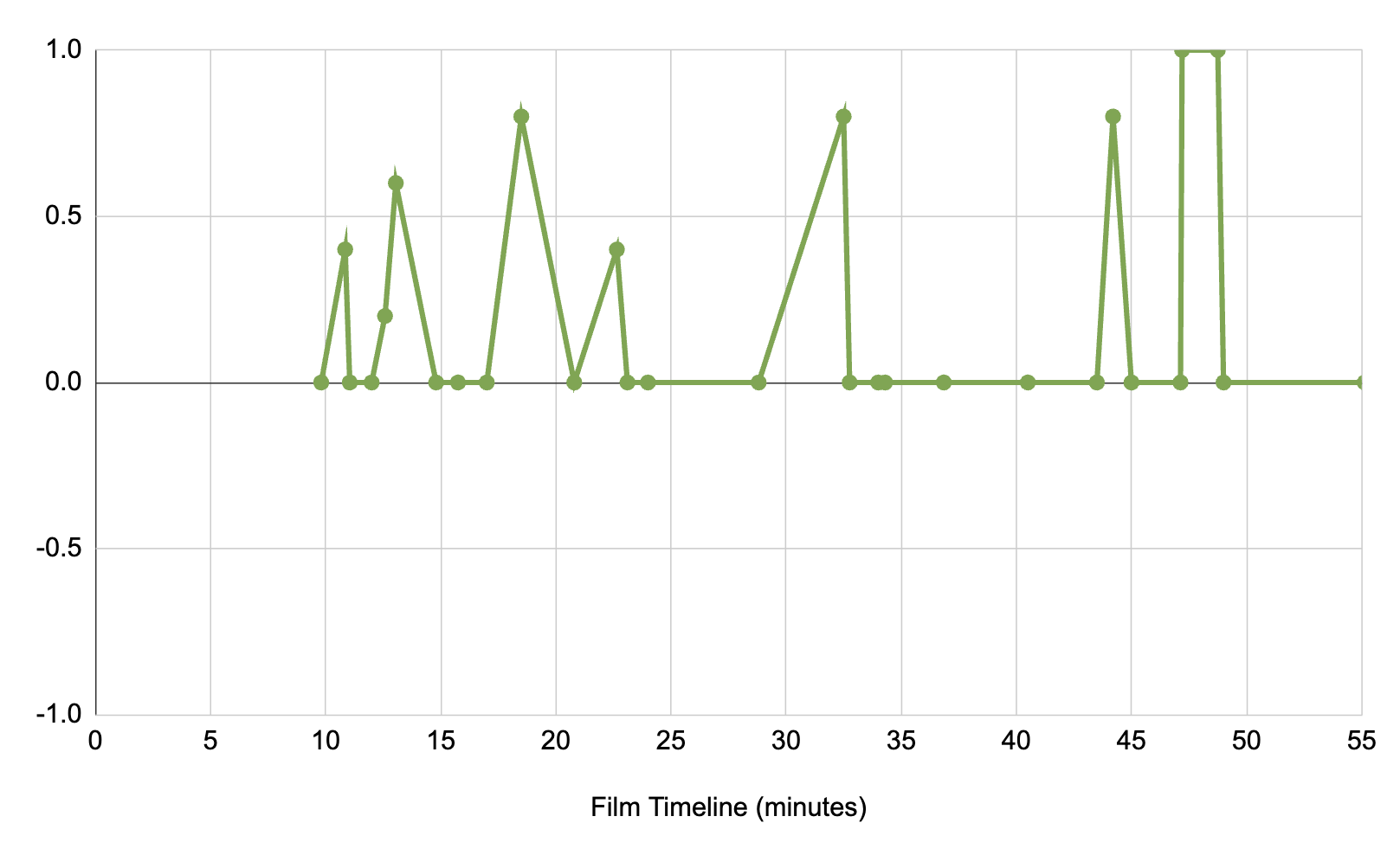}  
 \caption{This function provides the instances of moments of surprise for the first 55 minutes of the movie \emph{Psycho}. Note the varying magnitudes of the moments, as well as, the absence of Predictable moments (negative surprise).}
 \label{figure/images/Psycho_StoryMoments_A0_Suprise}
\end{figure}

The three types of universal Story moments are defined relative to the story, unlike Discourse moments which are defined relative to a specific character. Since, the three main types of evoked emotions toward a story can be considered three linearly independent axes between $1$ and $-1$, where one represents positive emotion, zero represents neutral emotion\footnote{Moments of Clarity with a 'neutral' value of 0 shall be considered a positively evoked emotion since only moments of Confusion (or negative clarity) will be recorded, see Section \ref{Simplifying Moments of Clarity} }, and minus one represents negative emotion, we are able to associate a numeric value with every instance of the three universal storytelling moments to indicate the perceived magnitude of that moment.  For entering these numbers, there is no need to be precise. However, their perceived relative importance needs to be captured. See Figure \ref{figure/images/Psycho_StoryMoments_A0_Suprise} for an example of varying magnitudes of surprise in the film \emph{Psycho}. 

%Figure \ref{figure/images/Psycho_StoryMoments_C0} provides the functions for the accumulated Story moments of the film \emph{Psycho}. Notice that Clarity does not contain negative values indicating there weren't any perceived moments of Confusion.    
%\begin{figure}[h] 
 %\includegraphics[width=0.5\textwidth]{images/Psycho_StoryMoments_C0}  
 %\caption{These three functions provide the accumulated Story moments for the first 55 minutes of the movies \emph{Psycho}. The three major evoked emotions toward a story: (Red) Curiosity, (Green) Surprise, and (Blue) Clarity.  The top function demonstrates that Alfred Hitchcock did not use any Predictable moments (negative surprise).}
% \label{figure/images/Psycho_StoryMoments_C0}
%\end{figure}

\subsection{Comparison of Audience Attachment to Story }
\begin{figure}[ht!]
 \includegraphics[width=0.5\textwidth]{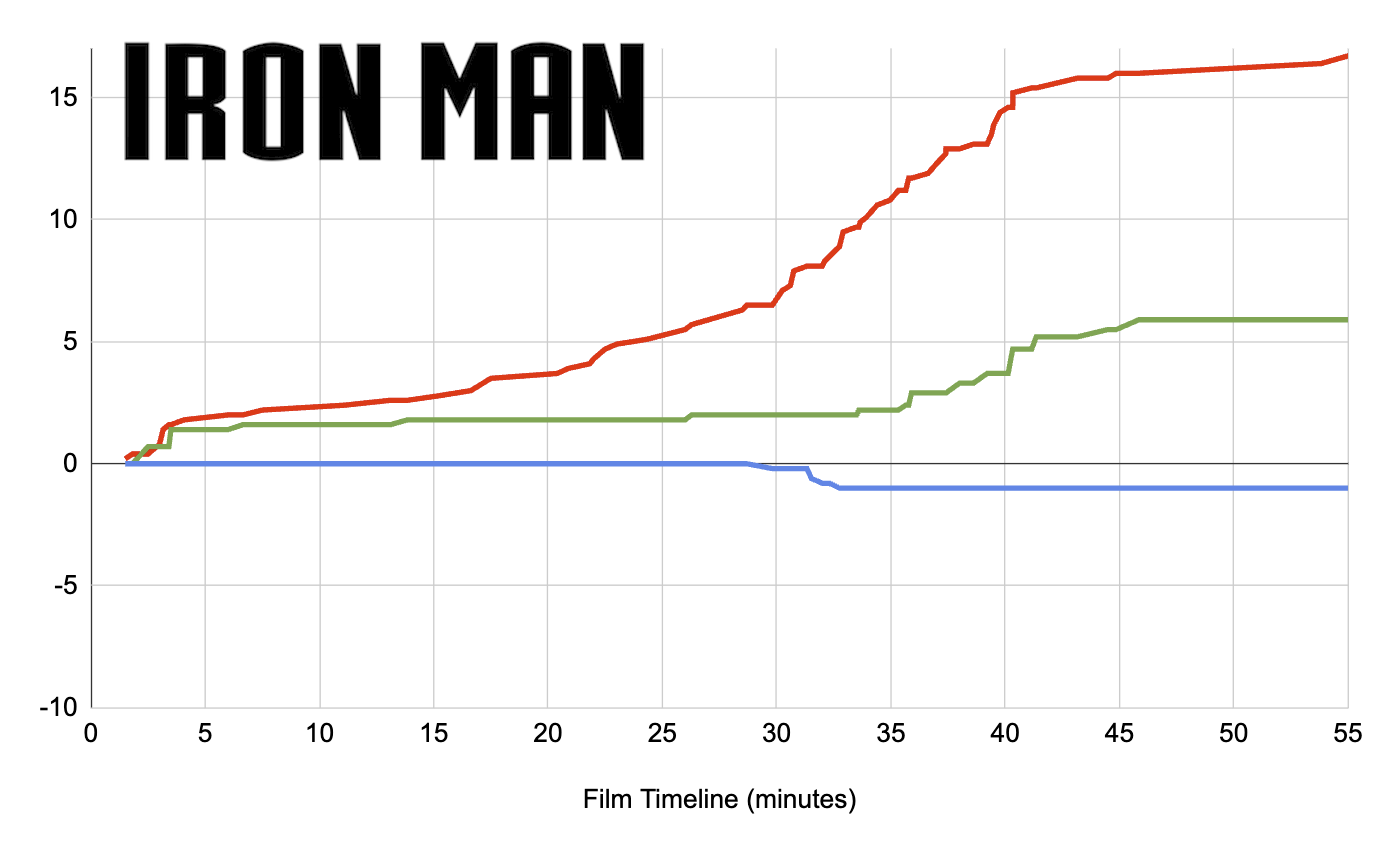}  
 \includegraphics[width=0.5\textwidth]{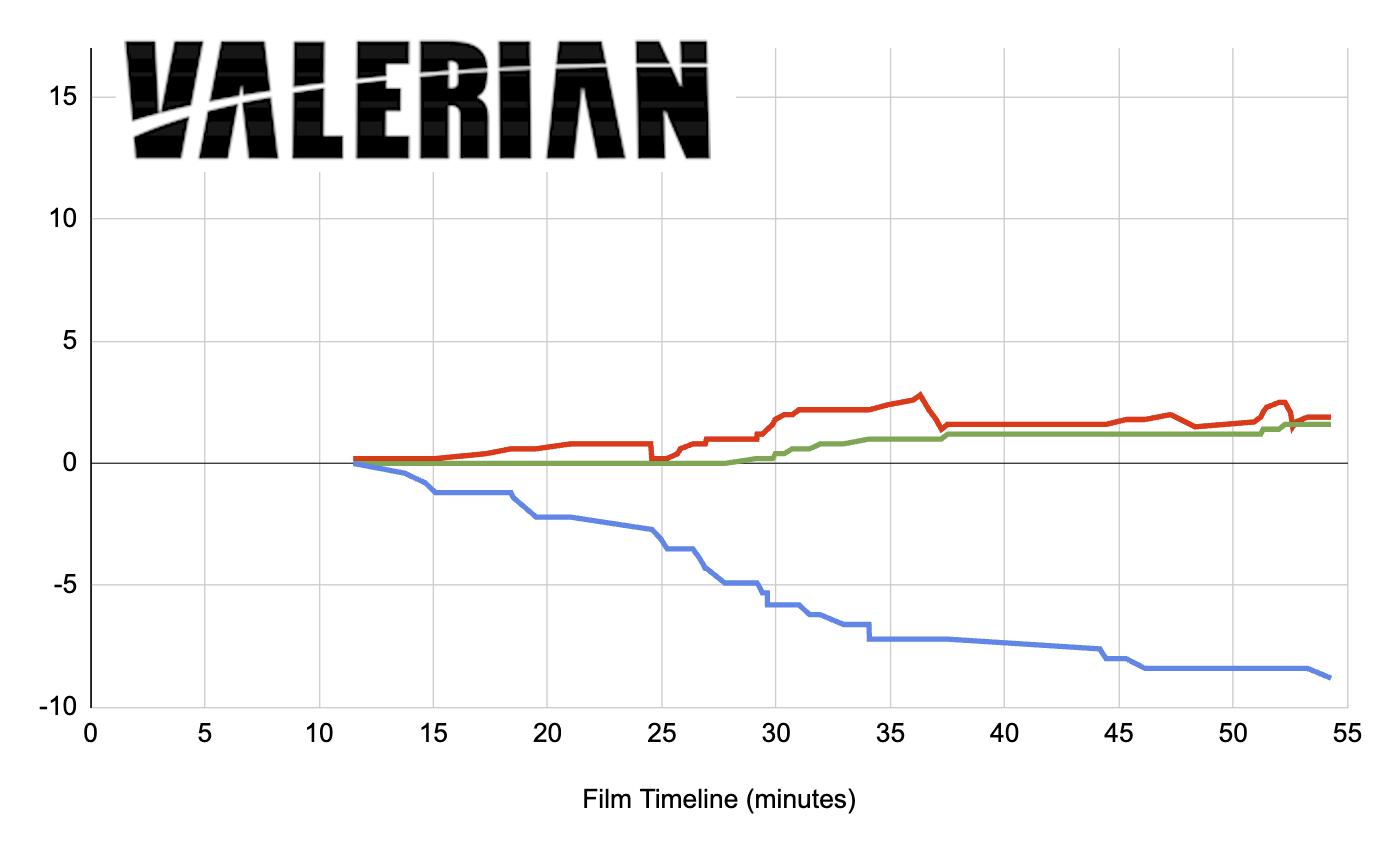}  
 \caption{Comparing the first 55 minutes of two Science-Fiction/Action/Adventure films: \emph{Iron Man} \& \emph{Valerian and the City of a Thousand Planets}
 featuring the three major evoked emotions toward a story: (Red) Curiosity, (Green) Surprise, and (Blue) Clarity. }
 \label{figure/images/IronManAndValerian}
 \end{figure}

\begin{figure}[ht!]
 \includegraphics[width=0.5\textwidth]{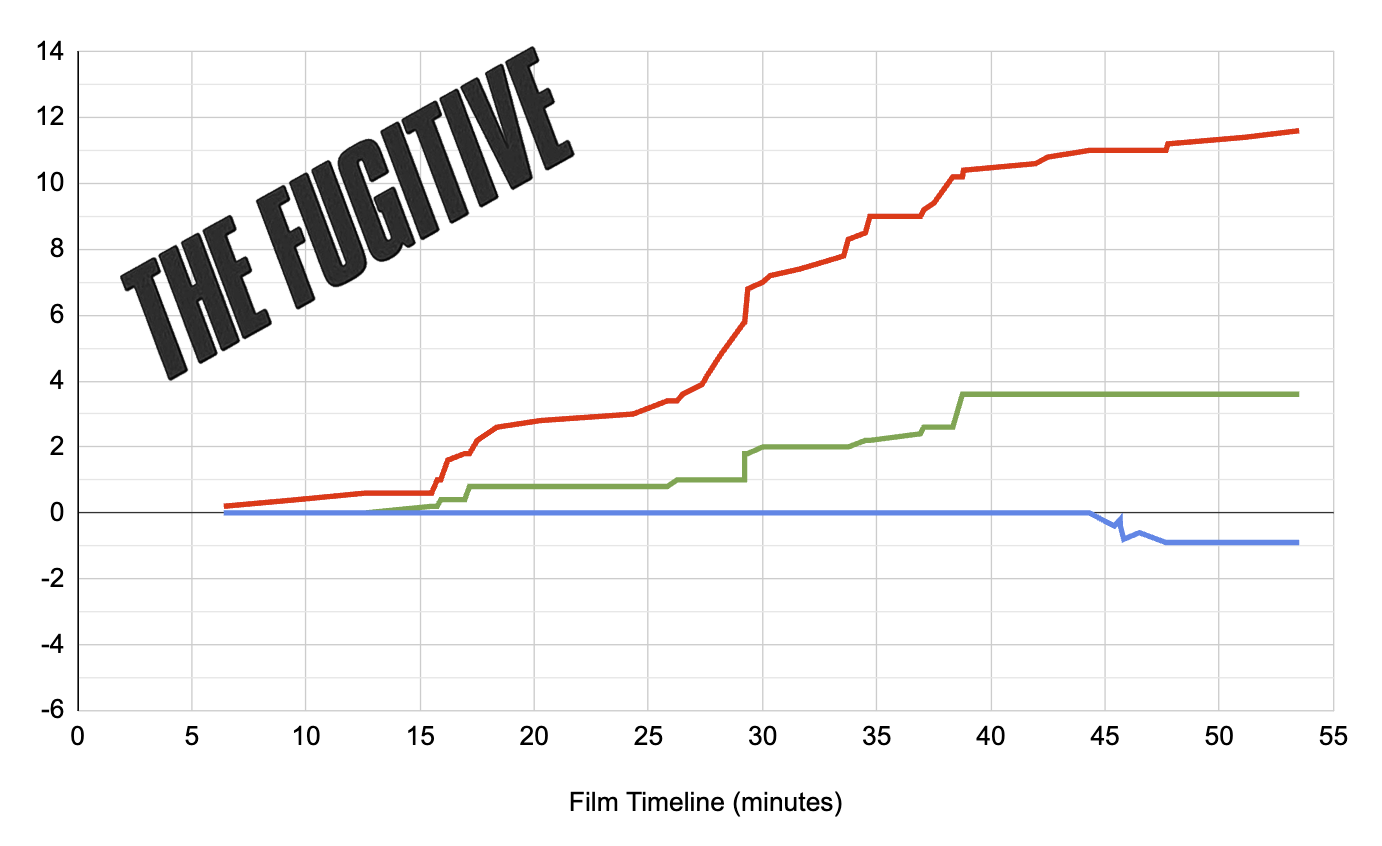}  
 \includegraphics[width=0.5\textwidth]{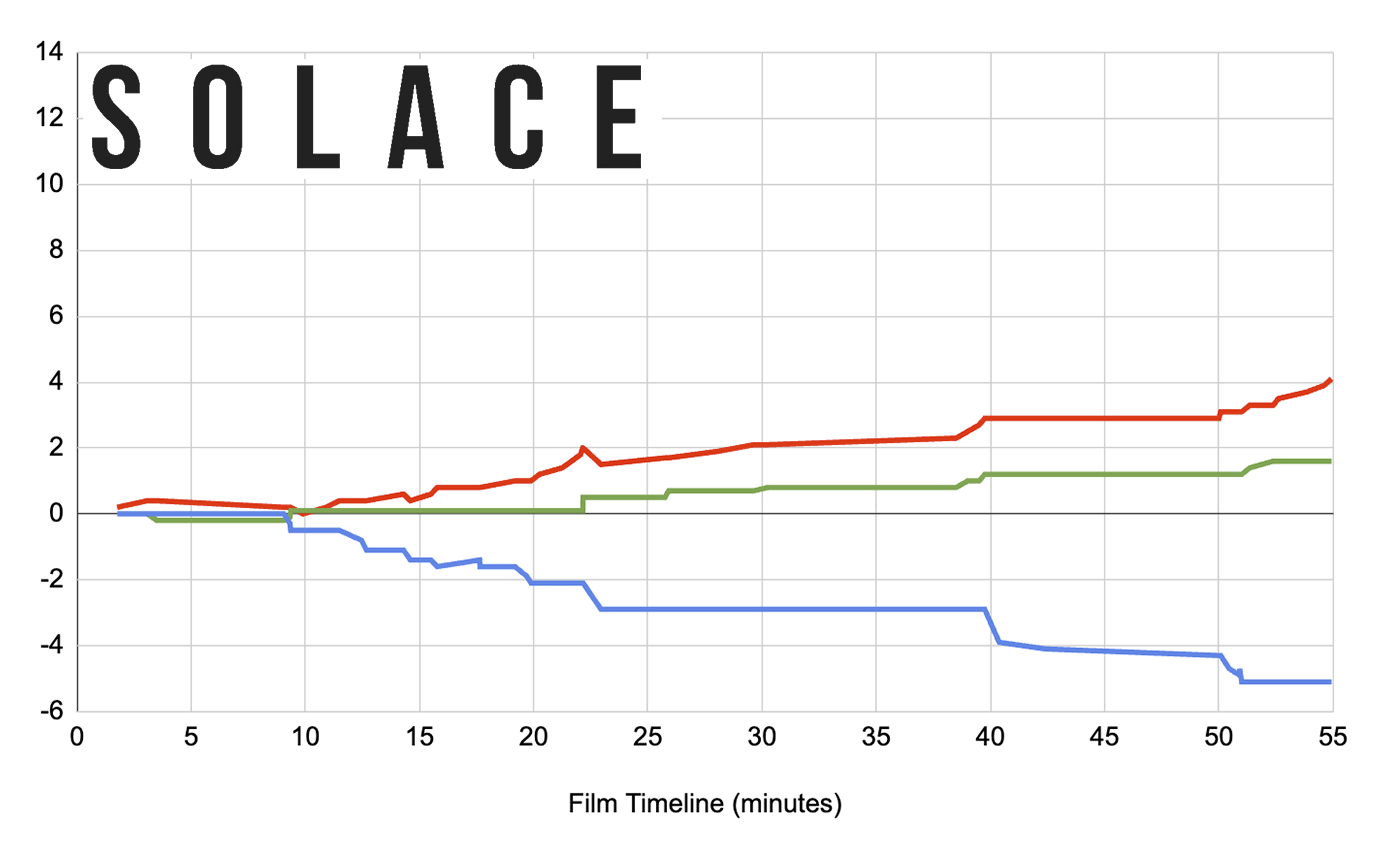}  
 \caption{Comparing the first 55 minutes of two Thriller/Mystery films: \emph{The Fugitive} \& \emph{Solace}
 featuring the three major evoked emotions toward a story: (Red) Curiosity, (Green) Surprise, and (Blue) Clarity. }
 \label{figure/images/TheFugitiveAndSolace}
 \end{figure}

Figure \ref{figure/images/IronManAndValerian} features the accumulated essential moments of Story for the two films of the same genre \emph{Iron Man} and \emph{Valerian and the City of a Thousand Planets} and similarly, Figure \ref{figure/images/TheFugitiveAndSolace} features two films of the same genre \emph{The Fugitive} (starring Harrison Ford) and \emph{Solace} (starring Anthony Hopkins).

Notice in \emph{Iron Man} and \emph{The Fugitive} the degree at which moments of Clarity dip into the negative territory especially compared to both \emph{Valerian} and \emph{Solace}. As a reminder, a neutral value for moments of Clarity corresponds with a positively evoked emotion, the lack of confusion, and negative values indicate the presence of confusion.

\section{Discussion}
\label{Sec/Discussion}

We think our analysis will particularly be useful in streaming movies. During streaming, the movies have to captivate their audience early in the film. Therefore, it is critical to attract the audience early in the movie. On the other hand, there is more leniency in theatrical movies since it is hard to leave the theater. Examples of such films are \emph{The Sixth Sense} and \emph{Red Sparrow}. Such movies should provide sufficient emotional attachments in discourse and story prior to their big surprise at the end otherwise people will be bored.

We observe that for audience attachment the neutral characters are the worst since the audience has no emotional response either positive or negative. With a negative accumulation of Discourse moments, the audience can feel despise or hatred for characters, which creates attachment. On the other hand, we also observe that a negative attachment to Story moments is not advisable since the audience can be bored. This can suggest that the values of Story moments should be between zero and one. 

In dealing with a limited range of -1 to 1 values for moments, the assigned numeric values are qualitative. In other words, they are meant only to be relative to other moments in the same film or relative to moments of other films. As far as each analyst is somewhat consistent and honest in their analysis, the overall analysis will be useful.
We think that it is even possible to avoid fine-grain analysis from the process and use the presence of a moment, which can be coded using only $1$ and $-1$. We expect that insight will still be available in such a coarse evaluation.

\subsection{Color Charts}
\label{Color Charts Section}
An additional implication of our representation of moments as vectors in $[-1,1]^3$ is that this cubical domain is isomorphic with RGB color space. Therefore, it can be visually represented using RGB colors. Let $\textbf{c}$ be an RGB color and $\textbf{c} \in [0,1]^3$, then we can convert $\textbf{m}$ to $\textbf{c}$, by the following rigid transformations: 
$$\textbf{c} = (\textbf{m} + \textbf{I})/2$$
where $\textbf{I} = (1,1,1)$. 

\begin{figure}[htb!]
\centering
    \begin{subfigure}[t]{0.35\textwidth}
        \includegraphics[width=1.0\textwidth]{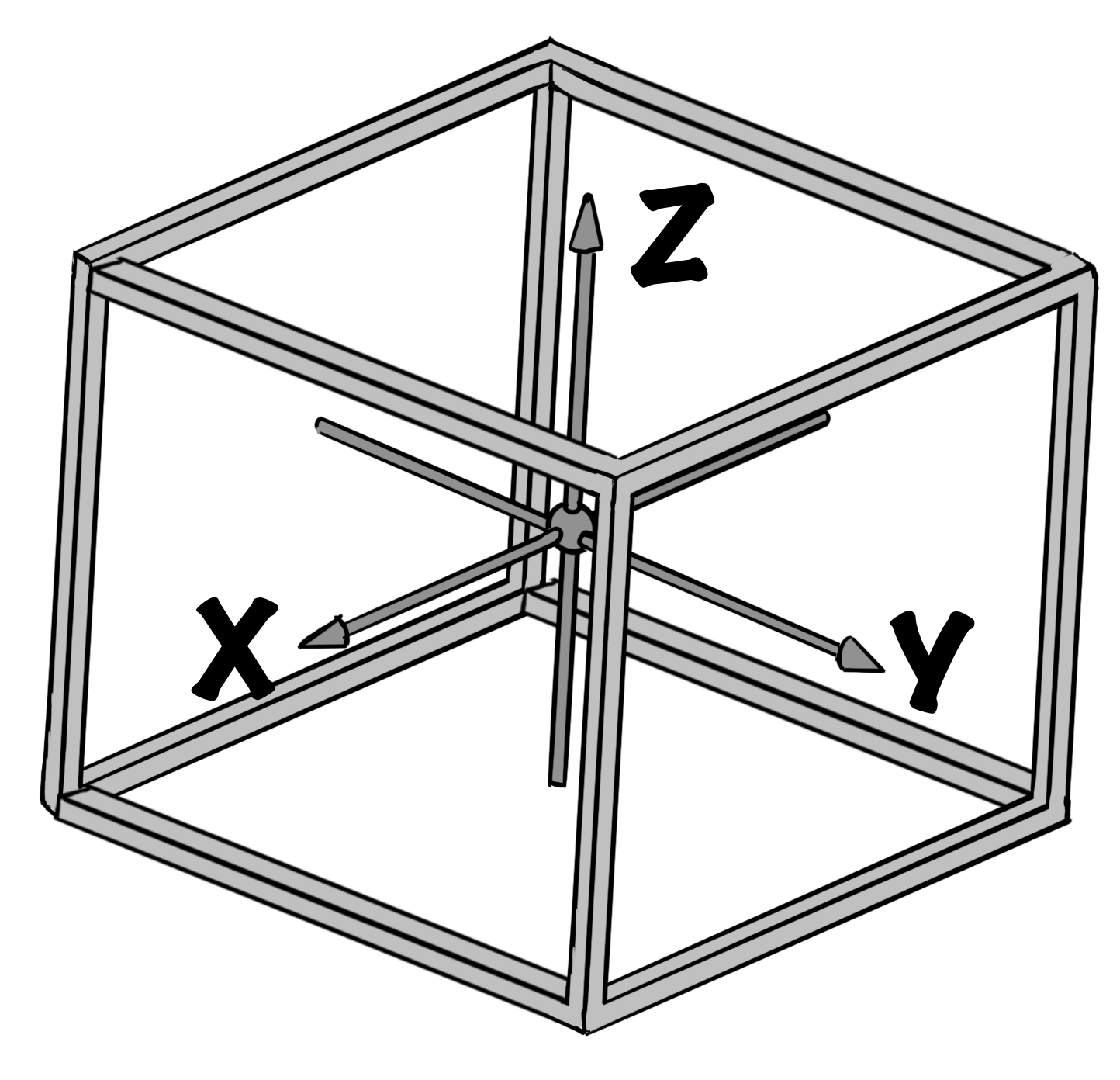}
        \caption{The domain of moments.}
        \label{figure/images/coordinates0}
    \end{subfigure}
        \begin{subfigure}[t]{0.35\textwidth}
        \includegraphics[width=1.0\textwidth]{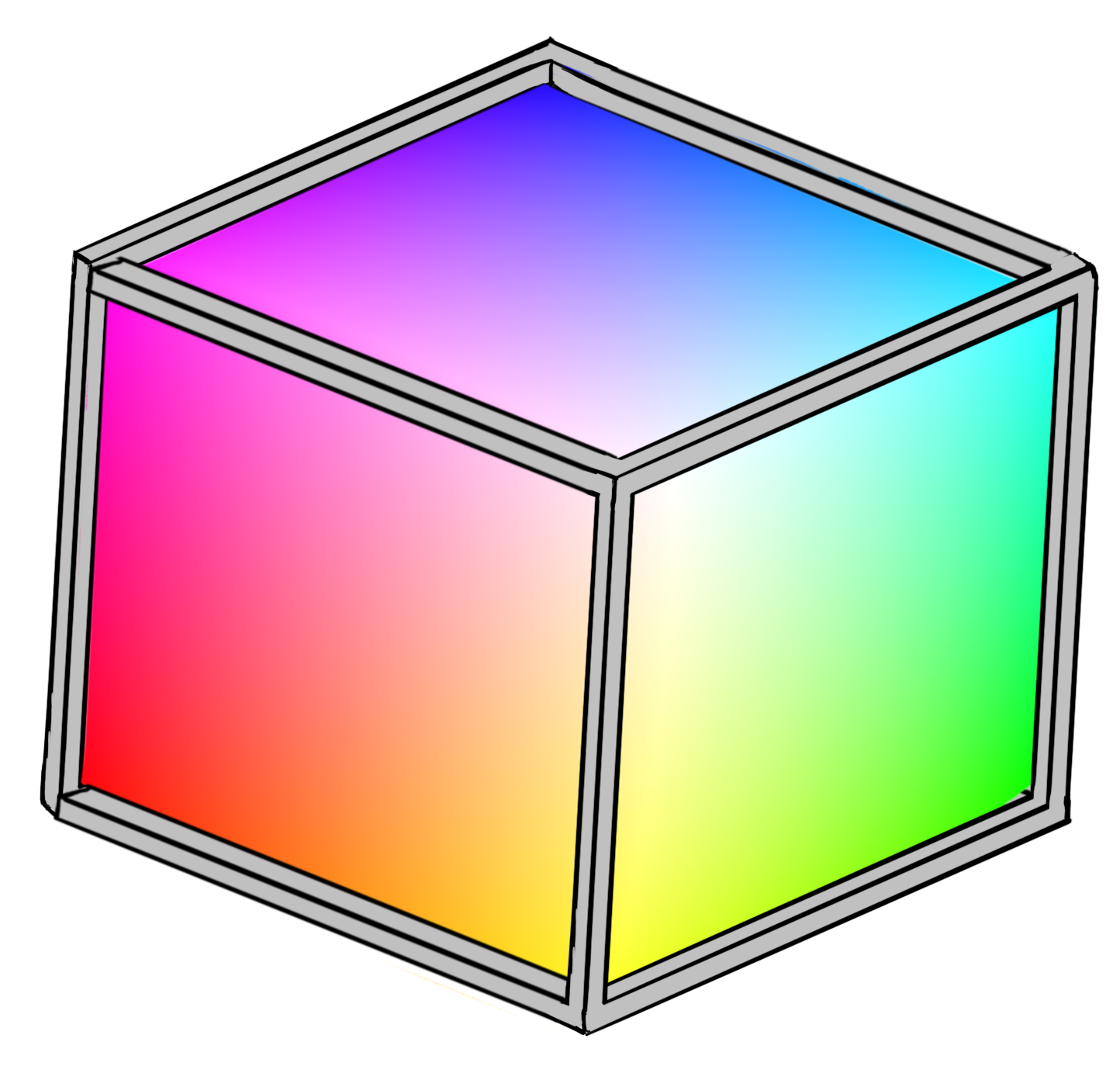} 
        \caption{RGB cube.}
        \label{figure/images/coordinates1}
    \end{subfigure}
\caption{This figure demonstrates that there is a simple isomorphic transformation from the domain of moments $[-1,1]^3$ to the domain of RGB colors $[0,1]^3$. }
\label{figure/images/coordinates}
\end{figure}

\begin{figure}[htb!]
\centering
         \includegraphics[width=0.8\textwidth]{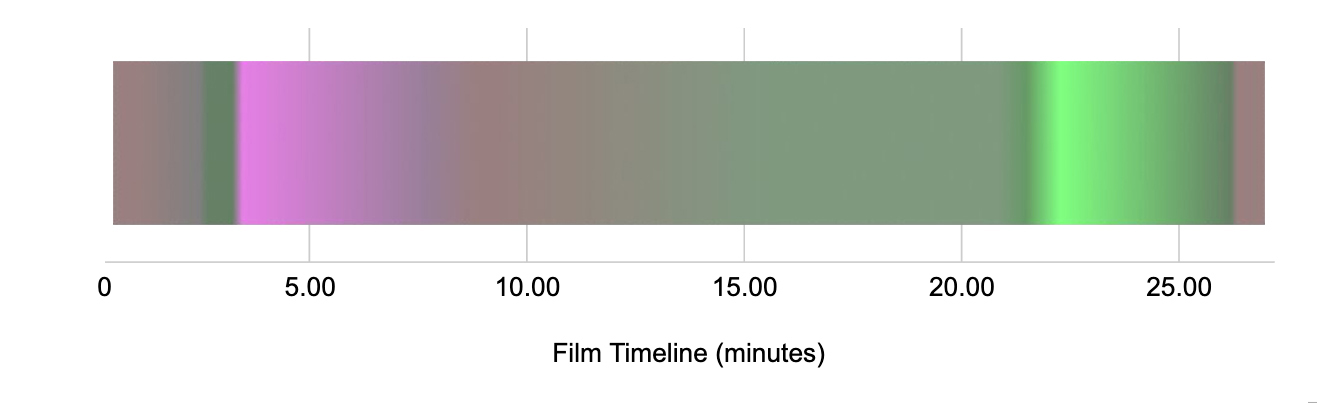}
\caption{This figure provides another visualization of discourse moments in Figure~\ref{figure/images/LadyBird_Marion_A0} as a colored chart where Red is Endearment, Green is Concern, and Blue is Justice. Note that this chart provides a good understanding of which types of moments are stronger in a given time.  }
\label{figure/images/LadyBird_Marion_A1}
\end{figure}

\begin{figure}[htb!]
\centering
         \includegraphics[width=0.8\textwidth]{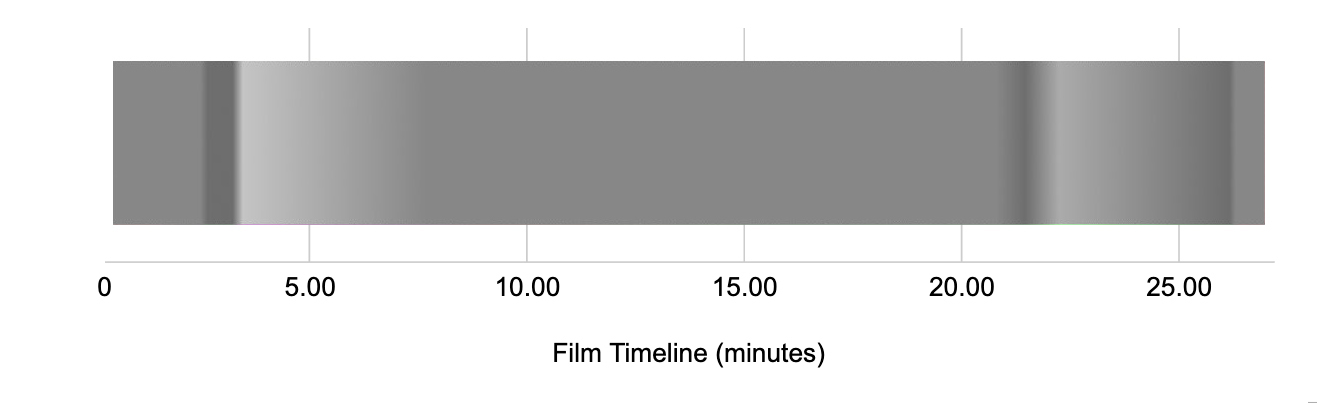}
\caption{This figure provides another visualization of a weighted average of discourse moments in Figure~\ref{figure/images/LadyBird_Marion_B0} as a Black and White chart. }
\label{figure/images/LadyBird_Marion_B1}
\end{figure}

This transformation also allows us to represent evoked emotion in a given time for a given character as a single color (See Figure~\ref{figure/images/coordinates}). If we create an image where the $x$ axis represents time steps and the $y$ axis represents different characters, we obtain an image and this image gives us another type of visual overview of a given visual story. Moreover, we can process this image using standard image processing to obtain more information. The authors of this paper prefer functions, but it could be possible that some people can read color charts better. 
One problem with color charts is that color-blind people may not be able to read color charts. One solution for color-blind is to convert RGB colors into a single-channel black-and-white image, which can represent only overall positive and negative emotions as attraction and repulsion. Figures~\ref{figure/images/LadyBird_Marion_A1}, and~\ref{figure/images/LadyBird_Marion_B1} shows color chart representations for 
Figures~\ref{figure/images/LadyBird_Marion_A0}, and~\ref{figure/images/LadyBird_Marion_B0} respectively. 

\subsection{Cumulative Moments as Color Charts}

\begin{figure}[thb!]
\centering
    \begin{subfigure}[t]{0.45\textwidth}
        \includegraphics[width=1.0\textwidth]{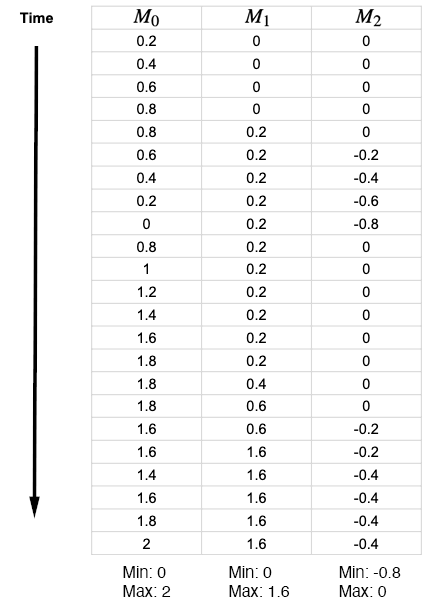}
        \caption{Cumulative Moments.}
        \label{figure/images/charts0}
    \end{subfigure}
    \begin{subfigure}[t]{0.45\textwidth}
        \includegraphics[width=1.0\textwidth]{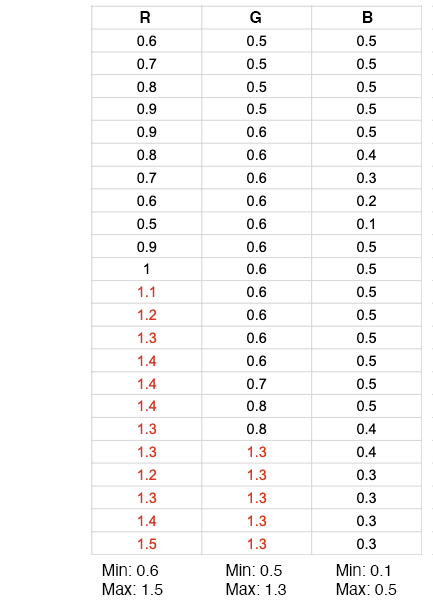}
        \caption{Conversion to RGB.}
        \label{figure/images/charts1}
    \end{subfigure}
    \caption{These two tables show accumulation data in moments and colors. Note that the numbers do not stay inside of the desired ranges. }
\label{figure/images/accumationtables}
\end{figure}

Note that the cumulative addition given in Equation~\ref{Equation/0} can go beyond the $[-1,1]^3$ box. This is clear in the examples in the table shown in Figure~\ref{figure/images/charts0}. This is not a problem for function-based visualization, however, it will be a problem if we represent these as color charts, since the corresponding color values also do not stay within the $[0,1]^3$ box (see the table in Figure~\ref{figure/images/charts1}). 

\begin{figure}[thb!]
\centering
        \begin{subfigure}[t]{0.45\textwidth}
        \includegraphics[width=1.0\textwidth]{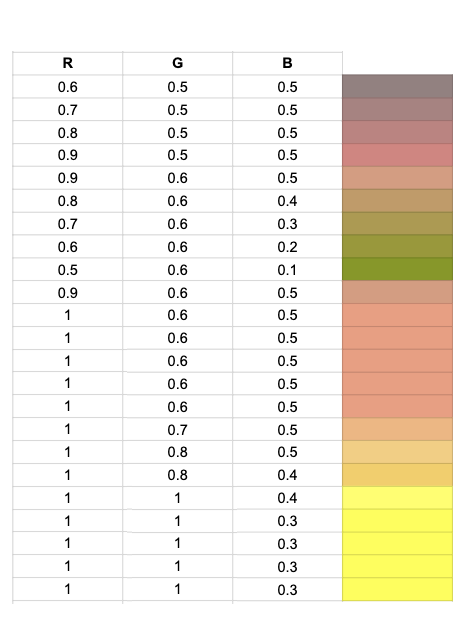} 
        \caption{Nonlinear Conversion to Color.}
        \label{figure/images/charts2a}
    \end{subfigure}
    \begin{subfigure}[t]{0.45\textwidth}
        \includegraphics[width=1.0\textwidth]{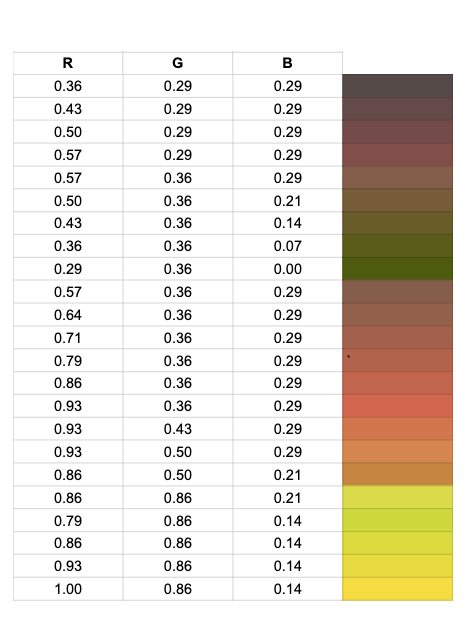} 
        \caption{Linear Conversion to Color.}
        \label{figure/images/charts2c}
    \end{subfigure}
\caption{This figure shows the effect of the different decisions affecting the final visualization with colors. This example is based on the discourse moments for the "Marion" character in the "Lady Bird" movie, which is widely analyzed in the literature \cite{stone2022lady, herdayanti2021psychological, musdalifa2022analysis}. }
\label{figure/images/charts}
\end{figure}

One solution is simply to truncate the values outside of the box. In other words, all the numbers larger than $1$ are truncated to $1$ and all numbers smaller than $0$ are truncated to $0$. This nonlinear operation creates the visualization shown in Figure~\ref{figure/images/charts2a}. 

\begin{figure}[htb!]
\centering
         \includegraphics[width=0.8\textwidth]{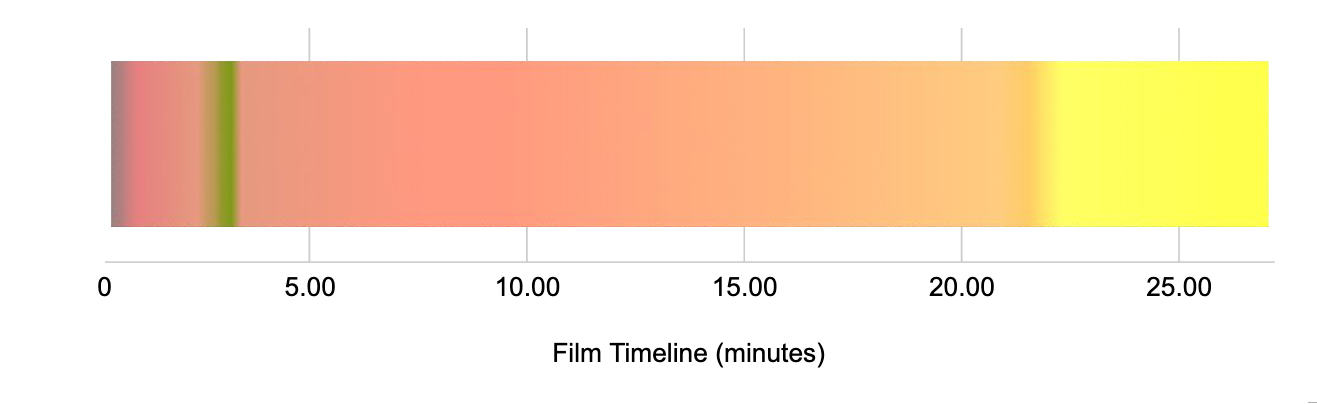}
\caption{This figure provides accumulated moments in Figure~\ref{figure/images/LadyBird_Marion_C0} as a colored chart where Red is Endearment, Green is Concern, and Blue is Justice. Notice the dominance of Concern (green) at around minute three. }
\label{figure/images/LadyBird_Marion_C1}
\end{figure}

Another option is to linearly transform the function in the range of $[-1,1]^3$ by using the following equation: 
$$M'_{k,2} = 2 \frac{M_{k,2} -M_{min}}{M_{max} - M_{min}} -1 $$
where $M_{max}$ and $M_{min}$ are maximum and minimum of all moments. Figure~\ref{figure/images/charts2c} shows the visual effect of this transformation. The problem with this transformation is that it is hard to compare charts of two characters visually. We always need to check actual numbers. On the other hand, the truncation in Figure~\ref{figure/images/charts2a} provides a consistent color scheme for visual comparison. Figure~\ref{figure/images/LadyBird_Marion_C1} shows the accumulated discourse moments in Figure~\ref{figure/images/LadyBird_Marion_C0} depicted along the film's timeline.

%\begin{figure}[htb!]
%         \includegraphics[width=0.5\textwidth]{images/LadyBird_Marion_D1}
%\caption{This figure provides another visualization of a weighted average of accumulated moments in %Figure~\ref{figure/images/LadyBird_Marion_D0} as a Black and White chart. }
%\label{figure/images/LadyBird_Marion_D1}
%\end{figure}

\subsection{Moments as Points}

 The moments can also be represented as points. In this paper, we do not use this representation, but it can be useful. Note that moments as points should be defined in homogenous coordinates as follows: 
$$ \textbf{m} = (T m_0, T m_1, T m2, T) = (m_0, m_1, m2, 1) $$
where $T$ is any positive number. The key idea here $T$ can be considered the time duration or the importance of this particular moment. This is useful since we can now compute a weighted average of moments. An additional advantage of representing moments as points is that the cumulative addition operations can produce only the elements of $[-1,1]^3$. Consider we have $K$ number of ordered moments $ \textbf{m}_k=(T_k m_{k.0}, T_k m_{k.1}, T_k m_{k.2}, T_k)$ where $k=1,2,\ldots,K$. If we add all these moments, we obtain the following formula: 
\begin{eqnarray}
\sum_{k=1}^{K} \textbf{m}_k &=& \left( \sum_{k=1}^{K} T_k m_{k,0}, \sum_{k=1}^{K} T_k m_{k,1}, \sum_{k=1}^{K}T_k m_{k,2}, \sum_{k=1}^{K}T_k \right) \nonumber \\
&=& \left( \frac{\sum_{k=1}^{K} T_k m_{k,0}}{\sum_{k=1}^{K}T_k}, \frac{\sum_{k=1}^{K} T_k m_{k,1}}{\sum_{k=1}^{K}T_k}, \frac{\sum_{k=1}^{K} T_k m_{k,2}}{\sum_{k=1}^{K}T_k}, 1 \right) \label{Equation/1}
\end{eqnarray}
Note that Equation~\ref{Equation/1} is a weighted average of all moments. Therefore, the resulting cumulative addition operation always stays inside of the $[-1,1]^3$  cube. 
Therefore, there is no need to re-scale the moments. Moreover, the cumulative values still be consistent with each other. Another advantage of this representation is that we can obtain color representations without any loss of information. On the other hand, we lose some of the convenience of vector representation. For instance, in this case, cumulative addition works like a cumulative average. The resulting curves would look different. We will not have other vector operations. It is, therefore, did not further investigate to use of moments as points. But, they could be useful in some applications. 

\section{Conclusion and Future Work}

The concept of moments is the main idea in this paper. We have identified all six moments by augmenting our observations with the existing publications in psychology and literature. Since there was no concept of the moment, we accessed the existing literature in not a completely systematic way. It is possible that we may have missed another type of moment that might also contribute to the audience's emotional attachment. If additional moment types are discovered, those new ones can be added to our model without a significant change. The model can potentially grow and be strengthened by such discoveries. 

Obtaining these functions with widespread data collection is almost impossible without the help of crowd-sourced efforts such as streaming companies that can collect the data from the audience during viewing. The same can be done while reading comic books on the internet. The effectiveness of our method can only be demonstrated by such widespread data collection that involves a wide range of audiences. 

Because of this underlying difficulty of testing our model, we view our approach as similar to physical theories. Initial theories are developed as thought experiments and they were later proved to be correct through observation, which was not possible when the theory was proposed first. We also want to point out that this is still not like a physical theory. Even with widespread data collection, the validity of our approach will only be based on statistical analysis. 

We also want to point out that there will be no universal truth similar to physical theories. A narrative structure that appears to be successful can go out of fashion with time. For instance, a surprising moment once learned can turn into predictable for most of the audience. Something that looks complex to one group of audience can look clear to another group. It is, therefore, important to correlate narrative structures with audience demographics, which will be dynamically changing. We, therefore, expect that this approach will also be helpful in tailoring movie selection based on the diversity of the audience. 

Another use of this approach is to identify new trends. For instance, some new and revolutionary narrative structures may start to be an attractive option for only a small group of niche audiences. These narrative structures can later be popular among widespread audiences. Identification of such revolutionary narrative structures early on can be helpful in deciding on diverging resources. 

\subsection{Limitations and Future Work}

The biggest limitation of our work, we ignore cause-and-effect relationships in favor of simplicity. In our framework, there is no genre or plot. We suggest looking at only cumulative emotions evoked by different moments. Such an approach cannot be successful 
in modeling or designing narrative structures. To create stories from scratch, we think that it is still better to formulate the problem using cause-and-effect relationships and focusing on genres and plots. On the other hand, our approach can be used to enhance the storytelling by adding moments. \emph{Psycho} is a good example. When we look carefully at the beginning of the movie we see that Alfred Hitchcock tried to capture the attention of the audience with a series of moments.   

One of the major limitations of our method is that it is hard to identify moments with some type of precision. Some people may simply overlook some moments. Therefore, we think cumulative moments are better since they tend to ignore high-frequency data and focus on only the trends, which can provide better intuition. 

We also want to point out that the interpolation formula is actually a first-degree non-uniform B-spline. We can replace it with higher degree B-splines to obtain smoother versions. Combining data can also be useful. These type of questions can only be answered when we have high-quality data collected by streaming companies or other crowd-sourced efforts.

\bibliographystyle{unsrtnat}
\bibliography{references}

\end{document}